\documentclass[letterpaper]{article} 
\usepackage{aaai2026}  
\usepackage{times}  
\usepackage{helvet}  
\usepackage{courier}  
\usepackage[hyphens]{url}  
\usepackage{graphicx} 
\usepackage{natbib}  
\usepackage{caption} 
\usepackage{algorithm}
\usepackage{algorithmic}
\usepackage{booktabs}
\usepackage{multirow}
\usepackage{xcolor}   
\usepackage[table,xcdraw]{xcolor}
\usepackage{subcaption}
\usepackage{amsmath,amsfonts}

\newcommand{\argmax}

\usepackage{newfloat}
\usepackage{listings}
\DeclareCaptionStyle{ruled}{labelfont=normalfont,labelsep=colon,strut=off} 
\floatstyle{ruled}
\newfloat{listing}{tb}{lst}{}
\floatname{listing}{Listing}
\title{Topological Federated Clustering via Gravitational Potential Fields under \\ Local Differential Privacy}
\author{
    Yunbo Long\equalcontrib\textsuperscript{\rm 1},
    Jiaquan Zhang\equalcontrib\textsuperscript{\rm 2,4},
    Xi Chen\textsuperscript{\rm 2,3}\thanks{Xi Chen is the corresponding author},
    Alexandra Brintrup\textsuperscript{\rm 1,5}
}
\affiliations{
    
    \textsuperscript{\rm 1}Department of Engineering, University of Cambridge\\
    \textsuperscript{\rm 2}Artificial Intelligence Innovation and Incubation Institute, Fudan University\\
    \textsuperscript{\rm 3}Shanghai Academy of AI for Science\\
    \textsuperscript{\rm 
    4}Shanghai Innovation Institute\\
     \textsuperscript{\rm 
    5}The Alan Turing Institute\\

}

\usepackage{bibentry}

\begin{document}

\maketitle

\begin{abstract}
Clustering non-independent and identically distributed (non-IID) data under local differential privacy (LDP) in federated settings presents a critical challenge: preserving privacy while maintaining accuracy without iterative communication. 
Existing one-shot methods rely on unstable pairwise centroid distances or neighborhood rankings, degrading severely under strong LDP noise and data heterogeneity. 
We present Gravitational Federated Clustering (GFC), a novel approach to privacy-preserving federated clustering that overcomes the limitations of distance-based methods under varying LDP.
Addressing the critical challenge of clustering non-IID data with diverse privacy guarantees, GFC transforms privatized client centroids into a global gravitational potential field where true cluster centers emerge as topologically persistent singularities. 
Our framework introduces two key innovations: (1) a client-side compactness-aware perturbation mechanism that encodes local cluster geometry as "mass" values, and (2) a server-side topological aggregation phase that extracts stable centroids through persistent homology analysis of the potential field's superlevel sets. 
Theoretically, we establish a closed-form bound between the privacy budget $\epsilon$ and centroid estimation error, proving the potential field's Lipschitz smoothing properties exponentially suppress noise in high-density regions.
Empirically, GFC outperforms state-of-the-art methods on ten benchmarks, especially under strong LDP constraints ($\epsilon < 1$), while maintaining comparable performance at lower privacy budgets. By reformulating federated clustering as a topological persistence problem in a synthetic physics-inspired space, GFC achieves unprecedented privacy-accuracy trade-offs without iterative communication, providing a new perspective for privacy-preserving distributed learning. The code is available at \url{https://github.com/Yunbo-max/Topological_Federated_Clustering}.

\end{abstract}


\section{Introduction}

Federated learning (FL) has emerged as a crucial privacy-preserving paradigm for training machine learning models on decentralized data \citep{McMahanMoore:FedAvg}.
Although initially developed for supervised tasks, FL is increasingly being adapted to unsupervised learning scenarios, particularly clustering, to analyze distributed unlabeled data.
Standard federated clustering methods, such as federated k-means \citep{DennisLiSmith:Heterogeneity}, exchange aggregated data representations (e.g., cluster centroids) rather than raw client data \citep{GarstReinders:FKM}. While this prevents direct exposure of individual data points, it fails to provide formal privacy guarantees. For example, a hospital with few pediatric cancer patients may publish a centroid so close to individual records that adversaries can reconstruct original diagnoses, e.g. in extreme conditions only one patient is recorded, hence the centroid is this individual's data.
Differential privacy (DP) provides rigorous privacy guarantees by injecting calibrated noise into computations. 
While centralized DP \citep{demelius2025recent} protects only the final aggregated output, local differential privacy (LDP) \citep{xia2020distributed} enforces privacy at the client level by requiring data randomization before any communication occurs. This is critical for federated learning scenarios where the server cannot be trusted. However, LDP poses significant challenges for clustering algorithms: the client-side noise required to achieve strong privacy ($\varepsilon < 1.0$) often disrupts the geometric relationships between points that clustering relies upon, leading to degraded utility. 

\begin{figure*}[t!]
    \centering
    \includegraphics[width=\textwidth]{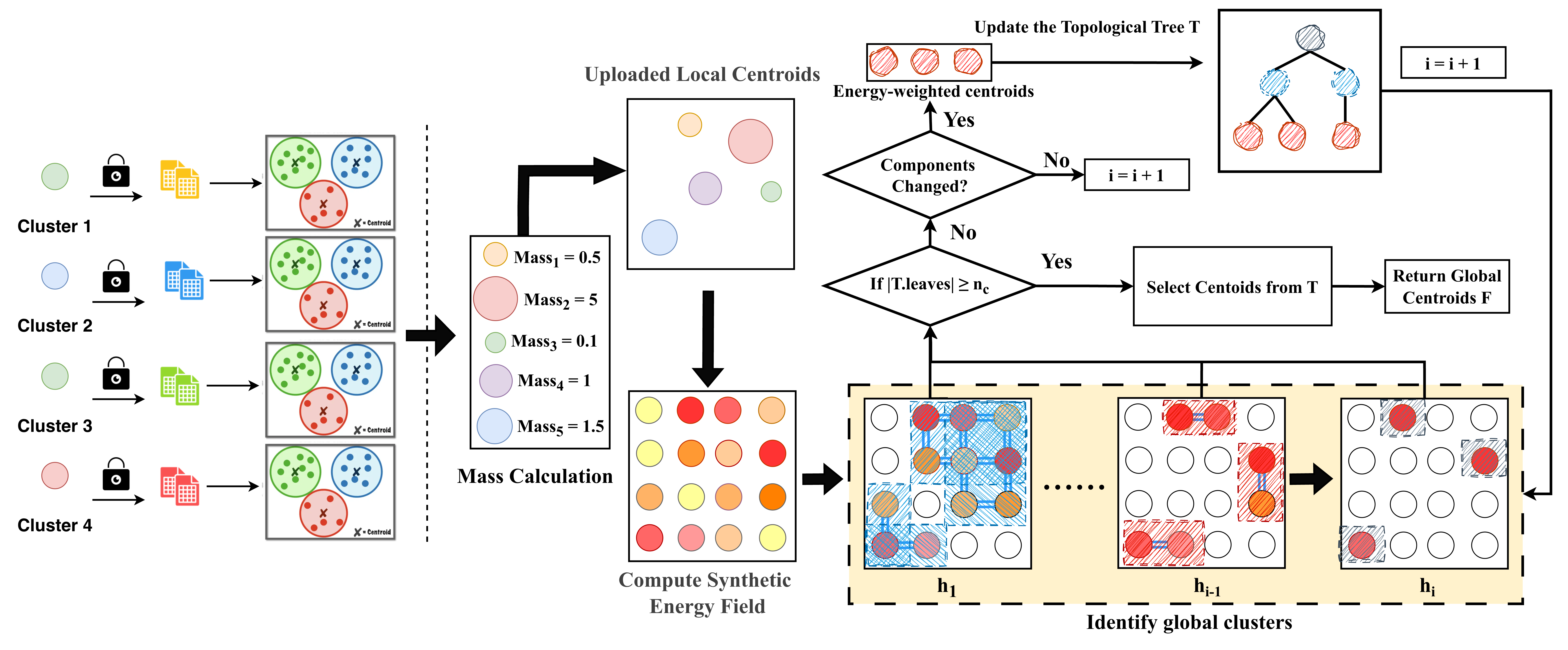}
    \caption{Gravitational Federated Clustering Pipeline.}
    \label{fig:GFC_framework}
\end{figure*}

Recently, FedDP-KMeans demonstrated the effectiveness of federated clustering under differential privacy by proposing an improved cluster initialization method combined with DP-Lloyds \citep{scott2025differentially}. However, its multi-round communication exacerbates privacy costs due to composition, as each Lloyd’s iteration requires fresh client-server interaction. 
This results in increased latency and bandwidth overhead. 
Such constraints explain the preference for one-shot methods in production FL systems, where communication efficiency is critical—especially given heterogeneous client connectivity. For instance, K-FED \citep{DennisLiSmith:Heterogeneity} clusters local centroids to derive global ones, while MUFC \citep{PanSima:Unlearning} enhances performance on imbalanced, non-IID data. 
The key challenge for one-shot federated clustering lies in accurately estimating global centroids under varying privacy budgets while limiting communication to a single round—eliminating the possibility of iterative refinement. Existing theoretical guarantees rely heavily on the assumption that geometric properties(connectedness and distance) remain intact after noise injection\citep{wang2024one}. However, this assumption breaks down in practical settings with strict privacy budgets (e.g., $\epsilon < 1$), where the noise magnitude becomes large enough to severely distort the underlying data structure, as evidenced by the catastrophic failure of baseline methods in Figure \ref{fig:varying_DP}.

Therefore, we propose Gravitational Federated Clustering (GFC), a one-shot federated clustering method designed to handle varying Local Differential Privacy (LDP) constraints. Unlike existing approaches that rely on noise-corrupted distance metrics or raw centroids, GFC reformulates clustering as a dynamic topological feature extraction process for global centroid search. By constructing a gravitational potential field from privatized client data (augmented with synthetic data), GFC robustly adapts to privacy budgets ranging from $\epsilon = 1000$ (weak privacy) to $\epsilon = 0.01$ (extremely strong privacy).
To summarize, our main contributions are as follows:
\begin{itemize}
    \item We propose the first one-shot federated clustering method GFC, that effectively handles varying LDP by modeling clustering as a topological persistence problem within a gravitational potential field. This avoids reliance on noise-sensitive distance metrics.

    \item We provide a theoretical analysis of topological feature extraction via homotopy equivalence and introduce an efficient tree-splitting algorithm to recover cluster structures from the potential field. The final centroids are then derived from the leaves of the constructed tree.

    \item Extensive experiments on ten real-world federated benchmarks show that GFC outperforms state-of-the-art one-shot methods under varying differential privacy (DP) constraints, particularly for small privacy budgets ($\varepsilon < 1$). We further present a \emph{privacy-accuracy boundary analysis} to quantify the relationship between centroid accuracy and the privacy budget.

\end{itemize}

\begin{figure*}[t]
    \centering
    \begin{minipage}[t]{0.26\textwidth}
        \centering
        \includegraphics[width=\linewidth]{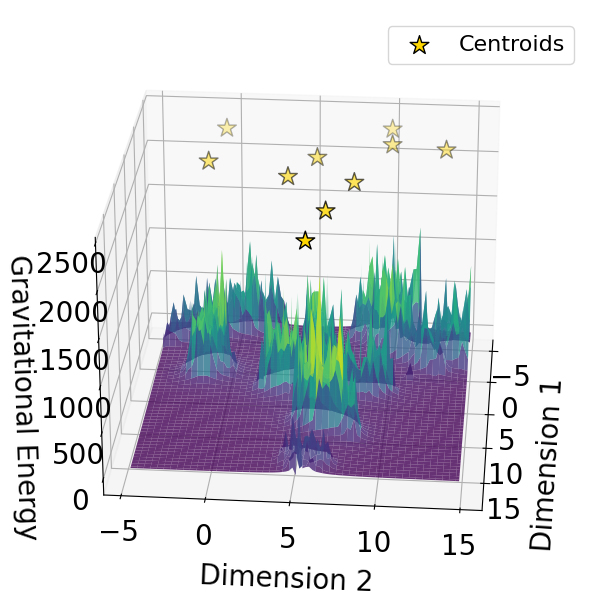}
        \par\vspace{0.1em}
        (a) 3D Plot
    \end{minipage}
    \hfill
    \begin{minipage}[t]{0.24\textwidth}
        \centering
        \includegraphics[width=\linewidth]{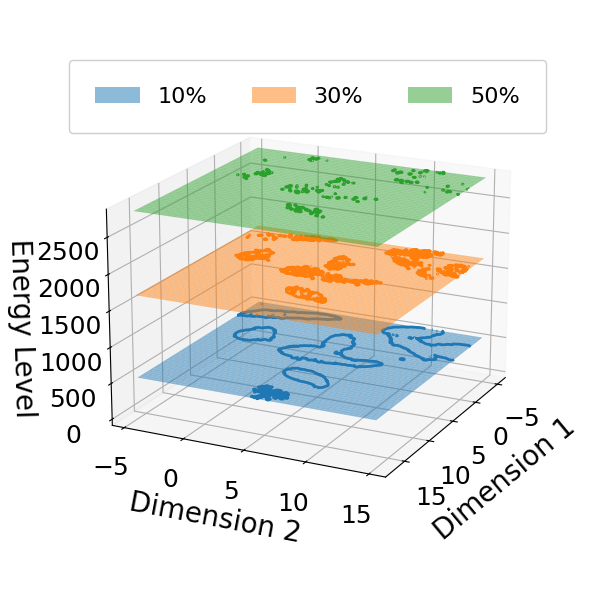}
        \par\vspace{0.1em}
        (b) Contours
    \end{minipage}
    \hfill
    \begin{minipage}[t]{0.24\textwidth}
        \centering
        \includegraphics[width=\linewidth]{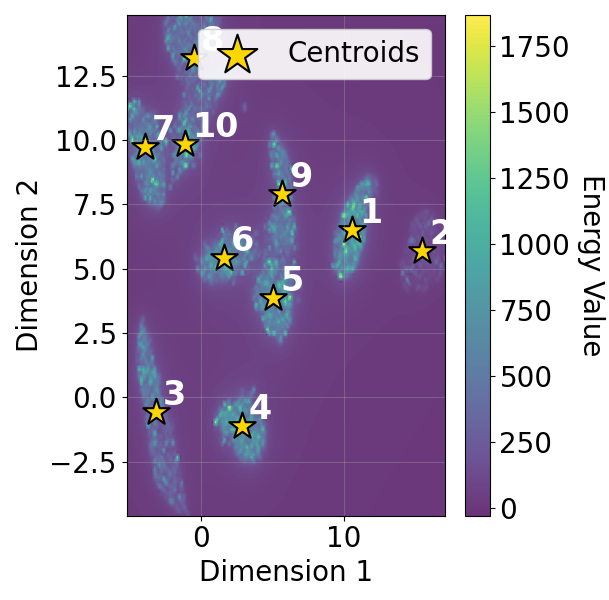}
        \par\vspace{0.1em}
        (c) 2D Gravitational Field
    \end{minipage}
    \hfill
    \begin{minipage}[t]{0.24\textwidth}
        \centering
        \includegraphics[width=\linewidth]{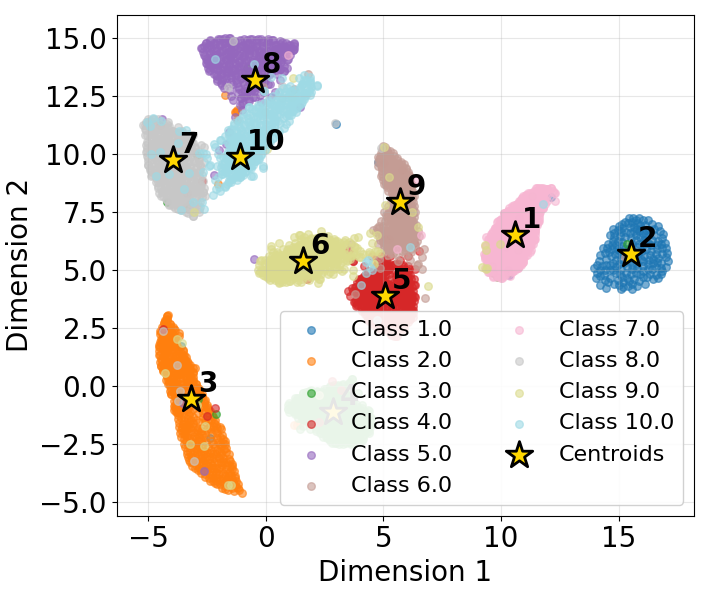}
        \par\vspace{0.1em}
        (d) Clustering
    \end{minipage}
    \caption{Examples of Topological Analysis for GFC on MNIST Data Visualized via UMAP Projection}
    \label{fig:topology}
\end{figure*}

\section{Related Work}
\label{sec:related_work}

Our work addresses the central challenge of achieving accurate clustering under Local Differential Privacy (LDP) in federated settings. 

\subsection{Federated Clustering}

Federated clustering enables decentralized data analysis by preserving privacy, avoiding raw data centralization. Early iterative methods like federated k-means required many communication rounds to refine centroids \citep{GarstReinders:FKM}. To enhance efficiency, one-shot methods became standard, where clients perform local clustering and send results (e.g., centroids or density cores) to the server for a single aggregation. Key examples include K-FED \citep{DennisLiSmith:Heterogeneity}, MUFC \citep{PanSima:Unlearning}, and NNFC \citep{chen2024one}, which use advanced aggregation to handle non-IID and imbalanced data. To address data heterogeneity, Clustered FL (CFL) groups clients by distribution to train specialized models \citep{SattlerMuller:CFL, GhoshChung:IFCA,long2025}, enhancing performance and fairness \citep{zhang2024enhancing, GuptaTarushi:pFClus}. However, these methods are vulnerable under strong privacy constraints.

\subsection{Differential Privacy in Federated Clustering}

In differential privacy, bounded sensitivity is enforced by clipping each data point to satisfy $|\mathbf{x}_i|_1 \leq \Delta$, where $\Delta$ controls the maximum influence of any single sample and $\epsilon$ determines the privacy-utility trade-off. Clients then release privatized data as $\tilde{\mathbf{x}}_i = \mathbf{x}_i + \boldsymbol{\eta}_i$, with $\boldsymbol{\eta}_i \sim \text{Laplace}(0, \Delta/\epsilon)$, and we refer to $\epsilon$-differential privacy as DP with noise $\eta \sim \text{Laplace}(0, 1/\epsilon)$. To ensure formal privacy guarantees in federated clustering with an untrusted server, we adopt client-level local differential privacy (LDP), which treats the entire client dataset $\mathcal{D}_m$ as the privacy unit, following~\citep{scott2025differentially}. Unlike data-point-level DP, which protects individual samples, client-level LDP safeguards the whole dataset without scaling noise per data point, simplifying sensitivity control. However, this stronger privacy model introduces a conflict with clustering accuracy: under small $\epsilon$, the injected noise dominates pairwise distances, scaling with $\mathcal{O}(d/\epsilon^2)$ and destroying the geometric structure necessary for meaningful clustering. Consequently, existing private federated clustering methods maintain accuracy only under weak privacy budgets ($\epsilon > 1$), offering insufficient protection for sensitive applications. 
Our work, Gravitational Federated Clustering (GFC), is designed specifically to overcome this accuracy collapse by creating a robust representation of data topology that is resilient to the high levels of noise required by strong LDP.  here amke this concise


\section{Gravitational Federated Clustering}

\label{sec:methodology}

We introduce Gravitational Federated Clustering (GFC), shown in Figure \ref{fig:GFC_framework}, a novel one-shot federated clustering framework designed for robustness against the noise introduced by differential privacy. Traditional federated clustering methods that rely on precise distance calculations or nearest-neighbour graphs are often destabilized by the perturbations required to protect data privacy. GFC overcomes this limitation by reframing the clustering problem through the lens of physics and topology. It models the decentralized, noisy client data as a system of point masses and identifies the true cluster centers by finding the most topologically significant features of the resulting gravitational potential field.

The algorithm operates in two main phases: \textbf{(1) client-side processing}, where local centroids are extracted under differential privacy, and \textbf{(2) server-side global aggregation}, where a continuous energy landscape is constructed and analyzed using tools from Topological Analysis to reveal the global cluster structure. See the details of the GFC implementation in Algorithm \ref{alg:gfc}.

\subsection{Client-Side Processing}
Each client in the federated system generates privacy-preserving local centroids without revealing its raw data. Let client $m$ possess a local dataset $\mathcal{D}_m = \{\mathbf{x}_i\}_{i=1}^{N_m}$, where $N_m$ is the size of client $m$'s dataset and each data point $\mathbf{x}_i \in \mathbb{R}^d$. 

To protect privacy, each client applies $\epsilon$-LDP at the record level, as shown in Algorithm \ref{alg:gfc} (Line 3). Calibrated Laplace noise is added to each data point $\mathbf{x}_i \in \mathcal{D}_m$ to produce a noisy dataset $\tilde{\mathcal{D}}_m$. The noise is scaled by $\Delta$, the $L_1$-sensitivity of the data. 

After privatizing the data, the client computes the "Local Centroids" by running k-means on the noisy dataset $\tilde{\mathcal{D}}_m$. The $i$-th client's $j$-th cluster is denoted $\mathcal{C}_{ij}$ with its centroid $\mathbf{c}_{ij}$. To characterize the cluster's density, we compute its mass:
\begin{equation}
    w_{ij} = \exp\left(-\frac{\sum_{\mathbf{x} \in \mathcal{C}_{ij}} \|\mathbf{x} - \mathbf{c}_{ij}\|^2}{2\sigma_i^2}\right)
    \label{eq:mass}
\end{equation}
where $\sigma_i^2$ is the variance of $\|\mathbf{x}_\mu - \mathbf{x}_\nu\|^2$ for distinct $\mathbf{x}_\mu, \mathbf{x}_\nu \in \mathcal{D}_i$. 

In a single communication round, each client transmits its set of local centroids and their corresponding masses, $(\mathbf{c}_{ij}, w_{ij})$, to the central server and take the union, denoted $\mathcal{S}$. After re-indexing, we obtain $\mathcal{S} = \{\mathbf{c}_\alpha, w_\alpha\}_\alpha$.



\subsection{Server-Side Global Aggregation}
Upon receiving the weighted local centroids from all $m$ clients, the server performs a topological analysis to identify the final $N$ global cluster centers.

\subsubsection{Gravitational Potential Field Construction}
The server first defines the bounding box of the received centroids. It then generates a synthetic dataset $G$ by \textbf{uniformly sampling $\alpha k$ points} from within these bounds. The potential energy field is then evaluated at each of these synthetic points $\mathbf{g}_j \in G$:
\begin{equation}
    E(\mathbf{g}_j) = \sum_{(\mathbf{c}_\alpha, w_\alpha) \in \mathcal{S}} \frac{w_\alpha}{\|\mathbf{c}_\alpha - \mathbf{g}_j\|^p + \delta},
    \label{eq:potential_field}
\end{equation}
where $\delta$ is a small constant to avoid division by zero and $p$ is a hyperparameter (typically $p=2$ for a Coulomb-like potential) that controls the field's decay. Note: For high-dimensional data, this field is typically constructed in a low-dimensional embedding (e.g., via UMAP) to mitigate computational cost.

\subsubsection{Gravitational Potential Field Construction}
The server constructs a potential energy field over the data space $\mathbb{R}^d$. The potential energy $E(y)$ at any point $y \in \mathbb{R}^d$ is defined by summing the contributions of each point mass:
\begin{equation}
    E(y) = \sum_{(c_\alpha,w_\alpha) \in \mathcal{S}} \frac{w_\alpha}{\|c_\alpha - y\|^2 + \delta},
    \label{eq:potential_field}
\end{equation}
where $\delta$ is a small constant to avoid division by zero

In this formulation, dense regions of client centroids (the point masses) create high-potential peaks in the energy landscape, with more massive (more compact) centroids contributing more significantly. These peaks are hypothesized to correspond to the true global cluster centers.

\begin{algorithm}[t!]
\caption{Gravitational Federated Clustering (GFC)}
\label{alg:gfc}
\textbf{Input}: Distributed datasets $D = \{D_1,\ldots,D_C\}$ across $C$ clients, Privacy budget $\epsilon$, Local clusters per client $k$, Target clusters $n_c$, Softening factor $\delta$, Synthetic multiplier $\alpha$, Local centroids' mass variance $\sigma$, Neighborhood radius $r$. \\
\textbf{Output}: Global centroids $F$
\begin{algorithmic}[1]
\STATE \textbf{Client Phase:}
\FOR{each client $c \in \{1,\ldots,C\}$}
    \STATE $\tilde{D}_c \gets D_c + \text{Lap}(0,\Delta/\epsilon)$ \COMMENT{LDP protection}
    \STATE $\{C_{c1},\ldots,C_{ck}\} \gets \text{KMeans}(\tilde{D}_c, k)$
    \FOR{each centroid $C_{ci}$}
        \STATE $w_{ci} \gets \exp(-\frac{\sum_{x\in C_{ci}}\|x-C_{ci}\|^2}{2\sigma^2})$ \COMMENT{Mass calculation}
    \ENDFOR
    \STATE Upload $\{(C_{ci},w_{ci})\}_{i=1}^{k}$ to server
\ENDFOR

\STATE \textbf{Server Phase:}
\STATE $\mathcal{S} \gets \bigcup_c \{(C_{ci},w_{ci})\}_{i=1}^{k}$ \COMMENT{Aggregate centroids}
\STATE Generate synthetic data $G$ with $\alpha \cdot |\mathcal{S}|$ data points within data bounds $\mathcal{B} = [\min(\mathcal{S}), \max(\mathcal{S})]$
\FOR{each $g_j \in G$}
    \STATE $E(g_j) \gets \sum_{(C_i,w_i)\in \mathcal{S}} \frac{w_i}{\|g_j-C_i\|^2 + \delta}$ \COMMENT{Potential field}
\ENDFOR
\STATE $H \gets \Phi(E(G), \alpha \cdot |\mathcal{S}|)$ \COMMENT{Threshold sequence}
\STATE $\mathcal{T}.\text{init}()$ \COMMENT{Initialize empty topological tree}
\FOR{each $h_i \in H$}
    \STATE $F_{h_i} \gets \{g_j | E(g_j) \leq h_i\}$ \COMMENT{Sub-level set at threshold $h_i$}
    \STATE $\mathcal{C}_{\text{new}} \gets \text{CC}(F_{h_i})$ \COMMENT{Connected components}
    \IF{$\mathcal{C}_{\text{new}} \neq \mathcal{T}.\text{leaves}()$}
        \STATE $\mathcal{T}.\text{add\_nodes}(\mathcal{C}_{\text{new}})$ \COMMENT{Update tree structure}
        \FOR{each new leaf $L \in \mathcal{T}.\text{new\_leaves}()$}
            \STATE $\mu_L \gets \frac{\sum_{g_j\in L}E(g_j)g_j}{\sum_{g_j\in L}E(g_j)}$ \COMMENT{Energy-weighted centroid}
            \STATE $\mathcal{T}.\text{store}(L, \mu_L)$ \COMMENT{Store cluster candidate}
        \ENDFOR
    \ENDIF
    \IF{$\lvert\mathcal{T}.\text{leaves}()\rvert \geq n_c$} 
        \STATE BREAK \COMMENT{Stop when sufficient clusters found}
    \ENDIF
\ENDFOR
\IF{$\lvert\mathcal{T}.\text{leaves}()\rvert < n_c$}
    \STATE $F \gets \mathcal{T}.\text{isolated\_paths}()$ \COMMENT{Get isolated components}
\ENDIF
\IF{$|F| < n_c$}
    \STATE $F \gets F \cup \mathcal{T}.\text{top\_energy\_leaves}(n_c-|F|)$ \COMMENT{Add high-energy leaves}
\ENDIF
\IF{$|F| < n_c$}
    \STATE Add top $n_c-|F|$ candidates by $E(g_j)$ \COMMENT{Final fallback}
\ENDIF
\STATE \textbf{return} $F$
\end{algorithmic}
\end{algorithm}

\subsubsection{Topological Feature Extraction}

To robustly identify cluster centers in the presence of differential privacy noise, we employ topological persistence analysis on the gravitational potential field. The method constructs a filtration through super-level sets $\mathcal{F}_h = \{\mathbf{g} \in G : E(\mathbf{g}) \geq h\}$, where $G$ represents the synthetic grid points and $h$ varies across energy thresholds. The function $\pi_0(\mathcal{F}_h)$, known as the \emph{zeroth homotopy group}, counts the number of \emph{path-connected components} in each super-level set. These components directly correspond to the distinct cluster peaks at energy level $h$.

The algorithm tracks these 0D-topological features (the components) as $h$ decreases in a process known as persistent homology. This allows us to distinguish stable, persistent peaks (true clusters) from minor, noisy peaks that merge quickly. The threshold sequence $H$ is generated by $\Phi(E(G), \alpha \cdot |\mathcal{S}|)$, scaling with data density through $\alpha \cdot |\mathcal{S}|$ where $|\mathcal{S}|$ is the total number of uploaded centroids. For each threshold $h_i \in H$, connected components are computed using the function $\text{CC}(F_{h_i})$, which constructs an adjacency matrix based on neighborhood radius $r$ and identifies connected components through graph traversal. For each persistent component $L$ identified in the merge tree, we compute an energy-weighted centroid $\mu_L = \frac{\sum_{\mathbf{g}\in L}E(\mathbf{g})\mathbf{g}}{\sum_{\mathbf{g}\in L}E(\mathbf{g})}$, naturally emphasizing regions of high potential energy that align with true cluster centers. The leaves of the resulting merge tree represent the most persistent peaks, which we identify as the final global cluster centers.
We employ a merge tree for efficient $O(m \log m)$ persistence computation, where $m$ is the number of synthetic points, avoiding costly recomputation at each threshold. If fewer than $n_c$ persistent peaks are found, a fallback mechanism hierarchically selects high-energy components followed by maximum-energy synthetic points to guarantee $n_c$ clusters. (See details in Appendix).

\section{Experimental Settings}
\label{sec:Experiments_settings}

\renewcommand{\arraystretch}{1.3} %
\begin{table*}[t]
\centering
\resizebox{\textwidth}{!}{%
\begin{tabular}{@{}lll*{8}{c}*{3}{c}@{}}
\toprule
\multirow{2}{*}{$\varepsilon$} & 
\multirow{2}{*}{\bf{Metric}} & 
\multirow{2}{*}{\bf{Method}} & 
\multicolumn{7}{c}{\textbf{Small to Medium-sized Datasets}} & 
\multicolumn{3}{c}{\textbf{Large Datasets}} \\ 
\cmidrule(lr){4-10} \cmidrule(l){11-13}
 & & & 
\textbf{Seeds} & \textbf{Thyroid} & \textbf{Breast} & \textbf{Heart} & \textbf{Gesture} & \textbf{Abalone} & \textbf{Waveform} & 
\textbf{Celltype} & \textbf{MNIST} & \textbf{Postures} \\ 
\midrule 
\multirow{10}{*}{\textbf{0.1}} &
  \multirow{5}{*}{\textbf{ARI}} &
  \textbf{K-Fed} &
  22.65±20.37 &
  \textcolor{blue}{\textbf{43.39±18.66}} &
  {1.89±3.69} &
  NA &

  0.24±0.81 &
  {11.16±2.61} &
  {30.40±6.69} &
  {17.32±2.24} &
  {39.56±4.19} &
  2.54±0.30 \\ \cline{3-13} 
 &
   &
  \textbf{MUFC} &
  {27.20±16.59} &
  42.85±17.24 &
  1.75±4.13 &
  {2.04±7.53} &
  {10.07±8.59} &
  8.01±5.27 &
  \textcolor{blue}{\textbf{33.80±7.80}} &
  17.27±2.27 &
  34.13±3.02 &
  {2.86±0.15} \\ \cline{3-13} 
 &
   &
  \textbf{NNFC} &
  1.36±5.67 &
  25.63±15.89 &
  0.67±2.48 &
  1.18±4.10 &
  3.92±8.33 &
  NA &
  5.14±7.25 &
  8.98±7.66 &
  8.23±8.30 &
  0.95±1.06 \\ \cline{3-13} 
 &
   &
  \textbf{GFC(Ours)} &
  \textcolor{blue}{\textbf{41.07±5.93}} &
  {43.38±9.93} &
  \textcolor{blue}{\textbf{2.19±3.92}} &
  \textcolor{blue}{\textbf{20.04±10.54}} &
  \textcolor{blue}{\textbf{13.49±11.06}} &
  \textcolor{blue}{\textbf{12.95±4.09}} &
  23.8±6.31 &
  \textcolor{blue}{\textbf{17.91±3.44}} &
  \textcolor{blue}{\textbf{56.53±5.95}} &
  \textcolor{blue}{\textbf{2.98±0.42}} \\ \cline{3-13} 
 &
   &
  \textbf{Improv.} &
  \cellcolor{gray!30}50.99\% $\uparrow$ &
  \cellcolor{gray!30}0.02\% $\downarrow$ &
  \cellcolor{gray!30}15.87\% $\uparrow$&
  \cellcolor{gray!30}882.35\% $\uparrow$ &

  \cellcolor{gray!30}33.96\% $\uparrow$ &
  \cellcolor{gray!30}16.04\% $\uparrow$ &
  \cellcolor{gray!30}29.58\% $\downarrow$ &
  \cellcolor{gray!30}3.41\% $\uparrow$&
  \cellcolor{gray!30}42.89\% $\uparrow$&
  \cellcolor{gray!30}4.20\% $\uparrow$ \\ 
  \cline{2-13} 
 & 
  \multirow{5}{*}{\textbf{NMI}} &
  \textbf{K-Fed} &
  28.84±22.16 &
  {39.40±12.00} &
  {1.61±2.42} &
  0.02±0.05 &
  0.78±2.06 &
  {11.88±1.82} &
  35.69±5.46 &
  {16.09±0.93} &
  {63.86±3.76} &
  {3.14±0.24} \\ \cline{3-13} 
 &
   &
  \textbf{MUFC} &
  {36.97±15.17} &
  38.64±10.89 &
  1.12±2.65 &
  \textbf{2.54±6.20} &
  {14.06±6.28} &
  9.32±4.34 &
  \textcolor{blue}{\textbf{40.21±4.41}} &
  15.91±1.10 &
  58.26±2.56 &
  3.12±0.34 \\ \cline{3-13} 
 &
   &
  \textbf{NNFC} &
  2.48±7.17 &
  25.16±13.94 &
  0.58±1.09 &
  1.19±3.40 &
  4.46±7.79 &
  NA &
  6.28±7.53 &
  9.02±7.26 &
  20.39±18.80 &
  1.38±1.06 \\ \cline{3-13} 
 &
   &
  \textbf{GFC(Ours)} &
  \textcolor{blue}{\textbf{46.09±8.40}} &
  \textcolor{blue}{\textbf{40.40±10.78}} &
  \textcolor{blue}{\textbf{1.67±3.09}} &
 
 \textcolor{blue}{\textbf{14.79±9.51}} &
  \textcolor{blue}{\textbf{17.47±6.14}} &
  \textcolor{blue}{\textbf{13.38±2.78}} &
  25.58±6.13 &
  \textcolor{blue}{\textbf{16.11±1.30}} &
  \textcolor{blue}{\textbf{72.25±3.91}} &
  \textcolor{blue}{\textbf{3.33±0.48}} \\ \cline{3-13} 
 &
   &
  \textbf{Improv.} &
  \cellcolor{gray!30}24.67\% $\uparrow$ &
  \cellcolor{gray!30}2.54\% $\uparrow$ &
  \cellcolor{gray!30}3.73\% $\uparrow$&
  \cellcolor{gray!30}482.28\% $\uparrow$ &

  \cellcolor{gray!30}24.25\% $\uparrow$ &
  \cellcolor{gray!30}12.63\% $\uparrow$ &
  \cellcolor{gray!30}36.38\% $\downarrow$ &
  \cellcolor{gray!30}1.24\% $\uparrow$&
  \cellcolor{gray!30}13.14\% $\uparrow$&
  \cellcolor{gray!30}6.05\% $\uparrow$ \\ \midrule 
\multirow{10}{*}{\textbf{0.05}} &
  \multirow{5}{*}{\textbf{ARI}} &
  \textbf{K-Fed} &
  8.74±16.04 &
  25.28±12.44 &
  {0.99±2.24} &
  0.04±0.16 &
  NA &
  {8.04±4.70} &
  {26.64±7.62} &
 {16.05±3.33} &
  {26.68±5.32} &
  2.30±0.60 \\ \cline{3-13} 
 &
   &
  \textbf{MUFC} &
  {18.05±20.36} &
  {26.62±16.22} &
  0.27±0.96 &
  {0.66±2.64} &
  {3.37±4.73} &
  7.20±3.63 &
  \textcolor{blue}{\textbf{28.94±5.60}} &
  14.64±3.15 &
  {26.78±5.80} &
  {2.85±0.22} \\ \cline{3-13} 
 &
   &
  \textbf{NNFC} &
  0.69±2.59 &
  23.10±11.95 &
  0.10±0.45 &
  NA &
  1.63±4.19 &
  NA &
  NA &
  0.87±3.65 &
  2.36±4.91 &
  NA \\ \cline{3-13} 
 &
   &
  \textbf{GFC(Ours)} &
  \textcolor{blue}{\textbf{36.96±8.13}} &
  \textcolor{blue}{\textbf{29.24±12.64}} &
  \textcolor{blue}{\textbf{1.79±4.22}} &
  \textcolor{blue}{\textbf{12.35±11.49}} &
  \textcolor{blue}{\textbf{7.66±9.53}} &
  \textcolor{blue}{\textbf{10.88±5.51}} &
  21.9±7.24 &
  \textcolor{blue}{\textbf{16.91±3.57}} &
  \textcolor{blue}{\textbf{56.23±5.99}} &
  \textcolor{blue}{\textbf{2.99±0.46}} \\ \cline{3-13} 
 &
   &
  \textbf{Improv.} &
  \cellcolor{gray!30}104.76\% $\uparrow$ &
  \cellcolor{gray!30}9.84\% $\uparrow$ &
  \cellcolor{gray!30}80.81\% $\uparrow$&
  \cellcolor{gray!30}1771.21\% $\uparrow$ &
  
  \cellcolor{gray!30}127.30\% $\uparrow$ &
  \cellcolor{gray!30}35.32\% $\uparrow$ &
  \cellcolor{gray!30}24.33\% $\downarrow$ &
  \cellcolor{gray!30}5.36\% $\uparrow$&
  \cellcolor{gray!30}109.97\% $\uparrow$&
  \cellcolor{gray!30}4.91\% $\uparrow$ \\  \cline{2-13} 
 &
  \multirow{5}{*}{\textbf{NMI}} &
  \textbf{K-Fed} &
  12.48±19.17 &
  28.91±8.80 &
  {0.75±1.41} &
  0.03±0.13 &
  0.34±1.11 &
  {9.88±3.61} &
  {29.73±6.06} &
{15.10±1.63} &
  53.20±3.32 &
  2.84±0.48 \\ \cline{3-13} 
 &
   &
  \textbf{MUFC} &
  {22.45±23.47} &
  {28.92±10.03} &
  0.58±1.63 &
  {0.82±2.39} &
  {5.80±5.84} &
  9.67±3.39 &
  \textcolor{blue}{\textbf{32.24±4.74}} &
  14.45±1.65 &
  {53.51±4.37} &
  {3.09±0.32} \\ \cline{3-13} 
 &
   &
  \textbf{NNFC} &
  2.18±6.39 &
  23.17±9.53 &
  0.14±0.60 &
  NA &
  2.02±4.27 &
  NA &
  0.02±0.03 &
  1.16±3.40 &
  6.23±12.56 &
  NA \\ \cline{3-13} 
 &
   &
  \textbf{GFC(Ours)} &
  \textcolor{blue}{\textbf{42.22±8.60}} &
  \textcolor{blue}{\textbf{31.75±9.40}} &
  \textcolor{blue}{\textbf{0.82±2.12}} &
  \textcolor{blue}{\textbf{11.21±10.15}} &
  \textcolor{blue}{\textbf{11.78±7.28}} &
  \textcolor{blue}{\textbf{11.45±4.05}} &
  19.86±5.98 &
  \textcolor{blue}{\textbf{15.78±2.03}} &
  \textcolor{blue}{\textbf{72.15±4.18}} &
  \textcolor{blue}{\textbf{3.22±0.54}} \\ \cline{3-13} 
 &
   &
  \textbf{Improv.} &
  \cellcolor{gray!30}88.06\% $\uparrow$ &
  \cellcolor{gray!30}9.79\% $\uparrow$ &
  \cellcolor{gray!30}9.33\% $\uparrow$&
  \cellcolor{gray!30}1511.12\% $\uparrow$ &

  \cellcolor{gray!30}103.10\% $\uparrow$ &
  \cellcolor{gray!30}15.89\% $\uparrow$ &
  \cellcolor{gray!30}38.40\% $\downarrow$ &
  \cellcolor{gray!30}4.50\% $\uparrow$&
  \cellcolor{gray!30}34.83\% $\uparrow$&
  \cellcolor{gray!30}4.20\% $\uparrow$ \\  \midrule 
\multirow{10}{*}{\textbf{0.01}} &
  \multirow{5}{*}{\textbf{ARI}} &
  \textbf{K-Fed} &
  {2.86±10.87} &
  {18.55±13.95} &
  NA &
  NA &
  0.12±0.53 &
  {2.92±5.05} &
  2.25±5.58 &
  13.61±3.60 &
  19.21±8.81 &
  1.97±0.82 \\ \cline{3-13} 
 &
   &
  \textbf{MUFC} &
  0.89±3.48 &
  14.34±12.23 &
  {0.03±0.13} &
  NA &
  {1.01±3.49} &
  1.69±3.81 &
  {3.12±6.22} &
  {14.38±3.75} &
  {19.34±8.57} &
  \textcolor{blue}{\textbf{2.85±0.54}} \\ \cline{3-13} 
 &
   &
  \textbf{NNFC} &
  NA &
  4.97±9.06 &
  NA &
  NA &
  NA &
  NA &
  NA &
  6.3±0.7 &
  NA &
  NA \\ \cline{3-13} 
 &
   &
  \textbf{GFC(Ours)} &
  \textcolor{blue}{\textbf{5.10±14.16}} &
  \textcolor{blue}{\textbf{18.92±13.84}} &
  \textcolor{blue}{\textbf{0.27±0.95}} &
  \textcolor{blue}{\textbf{0.22±0.95}} &
  \textcolor{blue}{\textbf{8.82±10.28}} &
  \textcolor{blue}{\textbf{7.82±6.48}} &
  \textcolor{blue}{\textbf{9.64±9.23}} &
  \textcolor{blue}{\textbf{14.54±4.23}} &
  \textcolor{blue}{\textbf{52.79±4.42}} &
\textbf{2.79±0.50} \\ \cline{3-13} 
 &
   &
  \textbf{Improv.} &
  \cellcolor{gray!30}78.32\% $\uparrow$ &
  \cellcolor{gray!30}1.99\% $\uparrow$ &
  \cellcolor{gray!30}800.00\% $\uparrow$&
  \cellcolor{gray!30}$\infty$ $\uparrow$ &

  \cellcolor{gray!30}773.27\% $\uparrow$ &
  \cellcolor{gray!30}167.81\% $\uparrow$ &
  \cellcolor{gray!30}208.97\% $\uparrow$ &
  \cellcolor{gray!30}1.11\% $\uparrow$&
  \cellcolor{gray!30}162.62\% $\uparrow$&
  \cellcolor{gray!30}2.10\% $\downarrow$ \\  \cline{2-13} 
 &
  \multirow{5}{*}{\textbf{NMI}} &
  \textbf{K-Fed} &
  {4.03±12.95} &
  {18.51±13.98} &
  NA &
  {0.04±0.17} &
  0.21±0.92 &
  {4.51±6.05} &
  3.50±5.84 &
  {14.57±1.60} &
  42.13±12.28 &
  2.58±0.69 \\ \cline{3-13} 
 &
   &
  \textbf{MUFC} &
  2.23±0.07 &
  13.84±10.63 &
  {0.01±0.05} &
  NA &

  {1.80±4.57} &
  2.79±4.09 &
  {4.59±6.37} &
  14.53±1.74 &
  {42.34±12.08} &
  {3.17±0.41} \\ \cline{3-13} 
 &
   &
  \textbf{NNFC} &
  NA &
  4.93±8.84 &
  NA &
  NA &
  0.31±1.36 &
  NA &
  NA &
  NA &
  NA &
  NA \\ \cline{3-13} 
 &
   &
  \textbf{GFC(Ours)} &
  \textcolor{blue}{\textbf{5.80±16.32}} &
  \textcolor{blue}{\textbf{19.14±13.38}} &
  \textcolor{blue}{\textbf{0.26±0.95}} &
  \textcolor{blue}{\textbf{0.63±0.76}} &
  \textcolor{blue}{\textbf{9.69±6.33}} &
  \textcolor{blue}{\textbf{9.57±5.52}} &
  \textcolor{blue}{\textbf{11.68±8.17}} &
  \textcolor{blue}{\textbf{15.27±3.43}} &
  \textcolor{blue}{\textbf{69.63±2.90}} &
  \textcolor{blue}{\textbf{3.20±0.46}} \\ \cline{3-13} 
 &
   &
  \textbf{Improv.} &
  \cellcolor{gray!30}43.92\% $\uparrow$ &
  \cellcolor{gray!30}3.40\% $\uparrow$ &
  \cellcolor{gray!30}2500.00\% $\uparrow$&
  \cellcolor{gray!30}1475.00\% $\uparrow$ &

  \cellcolor{gray!30}438.33\% $\uparrow$ &
  \cellcolor{gray!30}112.20\% $\uparrow$ &
  \cellcolor{gray!30}154.46\% $\uparrow$ &
  \cellcolor{gray!30}4.80\% $\uparrow$&
  \cellcolor{gray!30}64.5\% $\uparrow$&
  \cellcolor{gray!30}0.95\% $\uparrow$ \\  \midrule 
\end{tabular}%

}
\caption{Performance comparison of GFC, NN-FC, K-Fed, and MUFC using Adjusted Rand Index (ARI) and Normalized Mutual Information (NMI). Results, shown as mean $\pm$ standard deviation over 10 seeds, are evaluated under different privacy budgets $\varepsilon$, where a smaller $\varepsilon$ (e.g., 0.01) indicates stronger privacy. Higher values denote better performance; best results are highlighted in \textcolor{blue}{\textbf{blue}}.}
\label{tab:strong_DP}
\end{table*}

\renewcommand{\arraystretch}{1} %
\begin{table}[h]
\centering
\begin{tabular}{lrrr}
\toprule
\textbf{Dataset} & \textbf{Samples} & \textbf{Dimensions} & \textbf{Clusters} \\ 
\midrule
Postures & 74,975 & 15 & 5 \\
MNIST & 70,000 & 784 & 10 \\
Celltype & 12,009 & 10 & 4 \\
Waveform & 5,000 & 21 & 3 \\
Abalone & 4,177 & 8 & 3 \\
Gesture & 1,747 & 18 & 5 \\
Heart & 303 & 13 & 2 \\
Breast & 277 & 9 & 2 \\
Thyroid & 215 & 5 & 3 \\
Seeds & 210 & 7 & 3 \\
\bottomrule
\end{tabular}
\caption{Diversity of real-world datasets used for benchmarking, ordered by the number of samples.}
\label{tab:full_datasets}
\end{table}

\begin{figure*}[t]
    \centering
    \begin{minipage}[t]{0.24\textwidth}
        \centering
        \includegraphics[width=\linewidth]{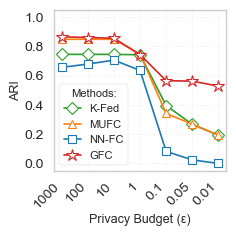}
        \par\vspace{0.1em}
        \textbf{(a) MNIST\textsubscript{ARI}}
    \end{minipage}
    \hfill
    \begin{minipage}[t]{0.24\textwidth}
        \centering
        \includegraphics[width=\linewidth]{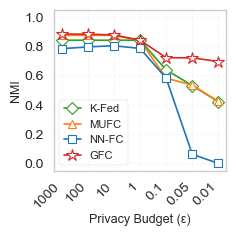}
        \par\vspace{0.1em}
        \textbf{(b) MNIST\textsubscript{NMI}}
    \end{minipage}
  \hfill
    \begin{minipage}[t]{0.24\textwidth}
        \centering
        \includegraphics[width=\linewidth]{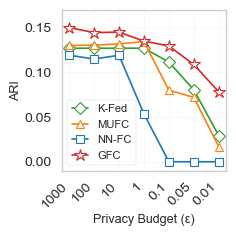}
        \par\vspace{0.1em}
        \textbf{(c) Abalone \textsubscript{ARI}}
    \end{minipage}
    \hfill
    \begin{minipage}[t]{0.24\textwidth}
        \centering
        \includegraphics[width=\linewidth]{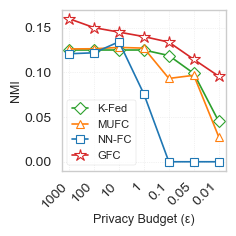}
        \par\vspace{0.1em}
        \textbf{(d) Abalone\textsubscript{NMI}}
    \end{minipage}
    \hfill
    \begin{minipage}[t]{0.24\textwidth}
        \centering
        \includegraphics[width=\linewidth]{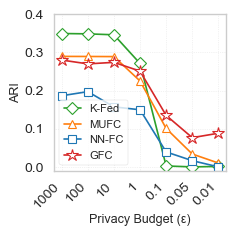}
        \par\vspace{0.1em}
        \textbf{(e) Gestures\textsubscript{ARI}}
    \end{minipage}
    \hfill
    \begin{minipage}[t]{0.24\textwidth}
        \centering
        \includegraphics[width=\linewidth]{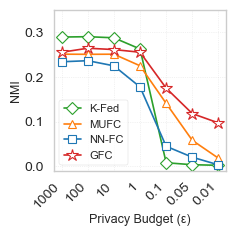}
        \par\vspace{0.1em}
        \textbf{(f) Gestures \textsubscript{NMI}}
    \end{minipage}
    \hfill
    \begin{minipage}[t]{0.24\textwidth}
        \centering
        \includegraphics[width=\linewidth]{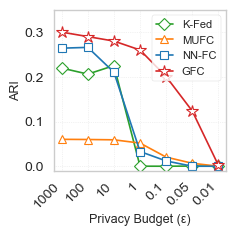}
        \par\vspace{0.1em}
        \textbf{(g) Heart \textsubscript{ARI}}
    \end{minipage}
    \hfill
    \begin{minipage}[t]{0.24\textwidth}
        \centering
        \includegraphics[width=\linewidth]{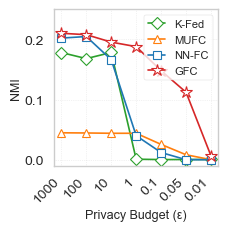}
        \par\vspace{0.1em}
        \textbf{(h) Heart\textsubscript{NMI}}
    \end{minipage}

    \caption{Performance comparison of GFC, NN-FC, K-Fed, and MUFC across varying privacy budgets $\varepsilon$ (smaller $\varepsilon$ denotes stronger privacy). Mean ARI scores are shown in (a, c, e, g), while corresponding mean NMI scores are in (b, d, f, h). GFC results are highlighted in red.}
    \label{fig:varying_DP}
\end{figure*}

\paragraph{\textbf{Non-IID Datasets Settings}}

We construct non-IID federated scenarios by partitioning datasets into \textit{num\_clients} subsets with heterogeneous label distributions. Our partitioning strategy applies $k$-means clustering followed by a systematic allocation process that ensures diverse label distributions across clients while maintaining balanced cluster representation. The detailed partitioning algorithm is provided in the Appendix.

\paragraph{\textbf{Baselines}}
We compare against three state-of-the-art one-shot federated clustering methods: k-FED \citep{dennis2021heterogeneity} for non-IID data handling, MUFC \citep{pan2022machine} for privacy through unlearning mechanisms, and NN-FC \citep{wang2024one} as the current SOTA using differential privacy and nearest-neighbor relationships. These cover key dimensions of accuracy, privacy, and heterogeneity adaptation.

\paragraph{\textbf{Experimental Setup and Evaluation}} 
We evaluate our GFC method against baselines using two robust clustering metrics: Adjusted Rand Index (ARI) \citep{steinley2004properties} for cluster similarity and Normalized Mutual Information (NMI) \citep{mcdaid2011normalized} for label agreement, accounting for label permutations and cluster size imbalances. All methods are tested under identical conditions across benchmark datasets, with 20 random seeds to assess consistency (reported as mean $\pm$ std). To analyze privacy-accuracy trade-offs, we fix $\Delta$ while varying $\epsilon \in [1000,100,10,1,0.1,0.05,0.01]$, measuring performance degradation as privacy strengthens.
Beyond assessing clustering performance (ARI/NMI) under varying privacy budgets, we conduct multi-dimensional analysis: (1) sensitivity to initial centroid count $k_0$ across dataset scales, (2) computational efficiency across diverse datasets, (3) ablation studies on gravitational parameters ($\delta \in [0.0001,0.1,100]$, synthetic data multiplier $\alpha \in [1,2,5,10]$) to quantify their impact on topological analysis, and (4) scalability tests with client numbers $C$ ranging from 10 to 100. 

\paragraph{\textbf{Hyperparameters Finetuning}} 

To ensure a fair comparison, all baseline methods utilize their originally published hyperparameters. For our method (GFC), we establish a systematic heuristic for hyperparameter selection grounded in the dataset's intrinsic properties and the allocated privacy budget $\epsilon$, which avoids extensive manual tuning. The number of local centroids $k$ scales with the dataset size $n$ as $k(n) = 15 + n/500$. The gravitational smoothing parameter is set to $\delta(\epsilon) = 500 \cdot e^{-5\epsilon}$, and the synthetic data multiplier follows $\alpha(\epsilon) = 2 + 20/(\epsilon + 1)$, ensuring both adapt to the privacy-utility trade-off. Finally, the neighborhood radius $r$ is dynamically determined as the $1^{\text{st}}$ percentile of non-zero distances between the received centroids, capturing the data's characteristic local scale. This principled approach, informed by persistent homology stability, allows for robust and reproducible configuration.

\section{Experimental Results}\label{sec:experiments_results}

\paragraph{Performance under strong LDP constraints.}
The results in Table \ref{tab:strong_DP} demonstrate that GFC consistently outperforms all state-of-the-art (SOTA) methods in accuracy, as measured by ARI and NMI, across datasets of varying sizes under privacy budgets ($\varepsilon$) ranging from 0.1 to 0.01.
For small datasets (e.g., Breast, Heart, and Gesture), GFC achieves significantly better performance under low privacy budgets, whereas other methods either fail (producing NA results) or yield near-zero accuracy. In medium and large datasets (e.g., Abalone and MNIST), GFC delivers substantial improvements: up to 167.81\% (Abalone) and 162.62\% (MNIST) in ARI under $\varepsilon = 0.01$ compared to baseline methods.
Furthermore, while existing SOTA methods exhibit high variance (reflecting instability under strong DP constraints), GFC maintains lower variance across most datasets, even as noise injection increases with stricter privacy budgets. 
On the high-dimensional Waveform dataset, GFC is competitive at moderate privacy and becomes the top performer at $\epsilon=0.01$, where other methods fail.
These results confirm GFC's accuracy and robustness in the federated clustering.

\paragraph{Performance under varying local DP.}

To evaluate the generalization of GFC under different privacy constraints, we test its clustering performance across a wide range of privacy budgets ($\varepsilon$), from 1000 (nearly non-private) to 0.01 (highly private), using both the Adjusted Rand Index (ARI) and Normalized Mutual Information (NMI) metrics. As illustrated in Figure \ref{fig:varying_DP}, GFC achieves comparable or even superior performance to baselines under high privacy budgets ($\varepsilon \geq 1$). As noise levels increase with stricter privacy constraints ($\varepsilon < 1$), GFC maintains relatively high accuracy with only a gradual decline, while competing methods suffer sudden performance drops. This advantage is particularly evident on large-scale datasets (e.g., MNIST) and small-to-medium datasets (e.g., Abalone and Heart), where GFC consistently outperforms state-of-the-art approaches across all privacy levels. These results highlight GFC's robustness under varying privacy constraints, making it a reliable choice for federated clustering. See further results in Appendix.

\paragraph{Ablation Study}

\paragraph{\textit{Sensitivity to local centroids number n}} 

We observed that under the privacy budget 0.1, GFC maintains stable performance across a wide range of local centroid counts (n), showing consistent accuracy on diverse datasets. Detailed results are listed in the appendix.

\paragraph{\textit{FC Efficiency.}} 

We compare the efficiency-accuracy trade-off between \textit{k}-Fed, MUFC, NN-FC, and our GFC framework (detailed in the appendix). While GFC exhibits comparable runtime performance to state-of-the-art (SOTA) methods—closely matching their speed on both small and large datasets—it achieves significantly higher clustering accuracy. Specifically, GFC maintains near-identical computational efficiency to k-fed, which is usually the fastest method (within 5\% time difference) but improves the average ARI by 167.81\% in the abalone dataset. This demonstrates that our energy-based formulation introduces comparable communication time cost while substantially enhancing federated clustering quality. Detailed runtime-accuracy curves are provided in the Appendix.

\paragraph{\textit{Number of synthetic data points.}} 

The number of synthetic data points ($n$) plays a crucial role in determining the energy field's approximation quality, with its impact varying significantly across privacy budgets. Our evaluation of $n \in \{1\times,2\times,10\times,100\times\}$ under $\epsilon \in \{1000,0.01\}$ (measured via ARI/NMI metrics in Table~\ref{tab:testing_charges}) reveals distinct patterns: For $\epsilon=1000$, small $n$ ($\leq1\times$) causes undersampling and unstable cluster boundaries, while moderate $n$ ($\approx2\times$ to $5\times$) achieves optimal balance between computational cost and field fidelity, and large $n$ ($10\times$) shows diminishing returns with increased communication overhead. However, under strict privacy ($\epsilon=0.01$), larger $n$ ($10\times$) becomes essential as high-energy centroids aggregate more closely, requiring denser sampling to maintain clustering accuracy - demonstrating how the optimal $n$ depends on both the dataset characteristics and privacy requirements.
More Results are shown in the Appendix.

\renewcommand{\arraystretch}{1.3} %
\begin{table}[t]
\centering
\begin{tabular}{cc|cccc}
\hline
\multirow{2}{*}{$\epsilon$} & \multirow{2}{*}{Metric} & \multicolumn{4}{c}{Number of Synthetic Data ($n$)} \\
 &  & 1 & 2 & 5 & 10 \\ \hline
\multirow{2}{*}{1000} 
 & ARI & 0.82 & 0.88 & 0.86 & 0.37 \\
 & NMI & 0.88 & 0.89 & 0.88 & 0.56 \\ \hline
\multirow{2}{*}{0.01} 
 & ARI & 0.10 & 0.11 & 0.15 & 0.53 \\
 & NMI & 0.34 & 0.35 & 0.34 & 0.70 \\ \hline
\end{tabular}
\vspace{0.2cm}
\\
\caption{Impact of Synthetic Data ($n$) and Privacy Budget ($\epsilon$) on Clustering Quality. Strong DP ($\epsilon=0.01$) requires a higher $n$ to compensate for noisier gradients.}
\label{tab:testing_charges}
\end{table}

\begin{figure}[h!]
    \centering
    \begin{minipage}[t]{0.15\textwidth}
        \centering
        \includegraphics[width=\linewidth]{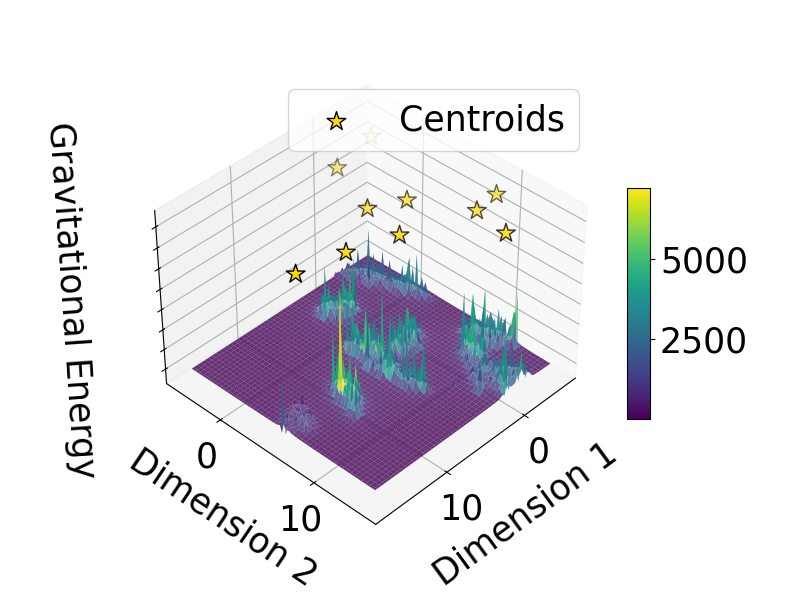}
        \par\vspace{0.1em}
        (a) delta=$1\times10^{-4}$
    \end{minipage}
    \hfill
    \begin{minipage}[t]{0.15\textwidth}
        \centering
        \includegraphics[width=\linewidth]{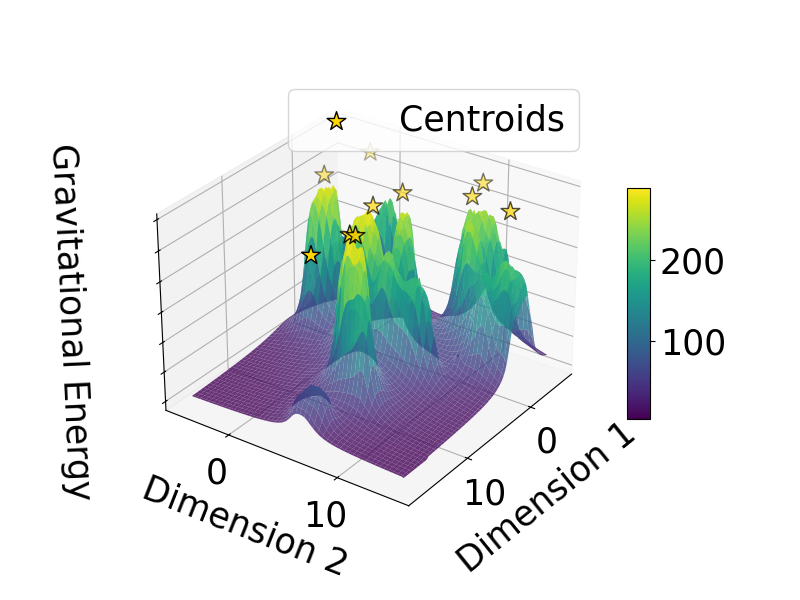}
        \par\vspace{0.1em}
        (b) delta=$1\times10^{-1}$
    \end{minipage}
    \hfill
    \begin{minipage}[t]{0.15\textwidth}
        \centering
        \includegraphics[width=\linewidth]{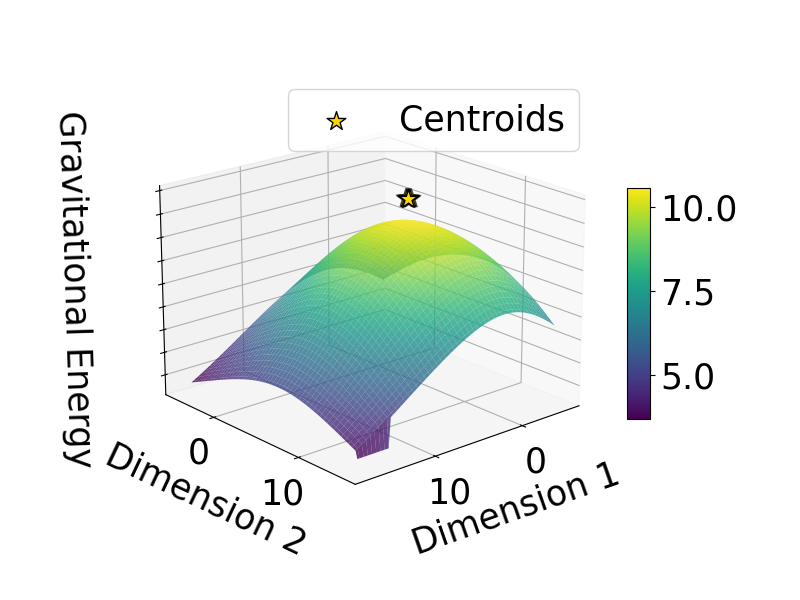}
        \par\vspace{0.1em}
        (c) delta=$1\times10^{2}$
    \end{minipage}
    \caption{Impact of $\delta$ for GFC on MNIST Data}
    \label{fig:energy_delta}
\end{figure}

\paragraph{\textit{Gravitational field construction.}} 
We analyze the effect of the smoothing parameter $\delta$ (tested at $1\times10^{-4}$, $1\times10^{-1}$, $1\times10^{2}$) on cluster formation and noise robustness in our potential field model. And $\delta$ controls the field's smoothness - smaller values create sharper energy field wells around synthetic points $\mathbf{s}_i$, while larger values produce smoother energy landscapes. Figure~\ref{fig:energy_delta} demonstrates this on MNIST clustering under low privacy regimes ($\epsilon=1000$), showing $\delta$'s impact on cluster separation and noise tolerance. Delta can be used to control the smoothness of the energy field under varying private budgets.
Extended results appear in the appendix.

\subsection{\textit{Scalability to Large Client Populations.}}

Our framework demonstrates strong scalability, maintaining stable and high clustering performance when extended to 100 or 1000 clients (see Appendix for full results). This scalability highlights our framework's practical applicability for real-world distributed systems with numerous participants.

\section{Discussion and Limitations}

Despite its strong performance, our framework has several limitations that point to future research directions. First, while our proposed heuristic methods automate much of the hyperparameter tuning, the model's performance can still be sensitive to these settings, particularly with high-dimensional datasets exhibiting highly non-IID client data distributions. Second, regarding scalability, although the algorithm is designed for a large number of clients, its communication overhead scales linearly with the client count. This becomes a significant bottleneck at an ultra-large scale (e.g., ten thousands of clients), as the server must aggregate information from every participant in each round.

\section{Conclusions}

We presented Gravitational Federated Clustering (GFC), a novel one-shot approach that reformulates private federated clustering as a topological persistence problem in a synthetic potential field. By encoding local cluster geometries as gravitational masses and extracting centroids through persistent homology analysis, GFC achieves: (1) robust performance across extreme privacy budgets ($\epsilon=0.01$ to $1000$), outperforming distance-based methods much on ARI and NMI accuracy under strong LDP ($\epsilon<1$); (2) provable noise suppression via the potential field's Lipschitz smoothing properties; and (3) elimination of iterative communication by proposing an one-shot method. 
The framework's client-side compactness-aware perturbation and server-side topological aggregation (Fig.~\ref{fig:GFC_framework}) provide a new paradigm for one-shot privacy-preserving Federated Clustering—particularly crucial for sensitive clustering applications like medical federations with strict local differential privacy requirements.

\section{Acknowledgment}

The computations in this research were performed using the CFFF platform of Fudan University

\bibliography{aaai2026}

@article{wang2024one,
  title={One-Shot Federated Clustering Based on Stable Distance Relationships},
  author={Wang, Yizhang and Pang, Wei and Pedrycz, Witold},
  journal={IEEE Transactions on Industrial Informatics},
  year={2024},
  publisher={IEEE}
}

@inproceedings{dennis2021heterogeneity,
  title={Heterogeneity for the win: One-shot federated clustering},
  author={Dennis, Don Kurian and Li, Tian and Smith, Virginia},
  booktitle={International conference on machine learning},
  pages={2611--2620},
  year={2021},
  organization={PMLR}
}

@article{pan2022machine,
  title={Machine unlearning of federated clusters},
  author={Pan, Chao and Sima, Jin and Prakash, Saurav and Rana, Vishal and Milenkovic, Olgica},
  journal={arXiv preprint arXiv:2210.16424},
  year={2022}
}

@inproceedings{McMahanMoore:FedAvg,
  title={Communication-Efficient Learning of Deep Networks from Decentralized Data},
  author={McMahan, H. Brendan and Moore, Eider and Ramage, Daniel and Hampson, Seth and y Arcas, Blaise Aguera},
  booktitle={Proceedings of the 20th International Conference on Artificial Intelligence and Statistics (AISTATS)},
  year={2017}
}

@article{SattlerMuller:CFL,
  title={Clustered Federated Learning: Model-Agnostic Distributed Multi-Task Optimization under Privacy Constraints},
  author={Sattler, Felix and M{\"u}ller, Klaus-Robert and Samek, Wojciech},
  journal={IEEE Transactions on Neural Networks and Learning Systems},
  year={2020}
}

@article{GhoshChung:IFCA,
  title={An Efficient Framework for Clustered Federated Learning},
  author={Ghosh, Avishek and Chung, Jichan and Yin, Dong and Ramchandran, Kannan},
  journal={arXiv preprint arXiv:2006.04088},
  year={2021}
}

@inproceedings{DennisLiSmith:Heterogeneity,
  title={Heterogeneity for the Win: One-Shot Federated Clustering},
  author={Dennis, Don Kurian and Li, Tian and Smith, Virginia},
  booktitle={International Conference on Machine Learning},
  pages={2611--2620},
  year={2021},
  organization={PMLR}
}

@article{GarstReinders:FKM,
  title={Federated K-Means Clustering},
  author={Garst, Swier and Reinders, Marcel},
  journal={arXiv preprint arXiv:2310.01195},
  year={2024}
}

@article{GuptaTarushi:pFClus,
  title={Fair Federated Data Clustering through Personalization: Bridging the Gap between Diverse Data Distributions},
  author={Gupta, Shivam and Tarushi and Wangzes, Tsering and Jain, Shweta},
  journal={arXiv preprint arXiv:2407.04302},
  year={2024}
}

@inproceedings{PanSima:Unlearning,
  title={Machine Unlearning of Federated Clusters},
  author={Pan, Chao and Sima, Jin and Prakash, Saurav and Rana, Vishal and Milenkovic, Olgica},
  booktitle={International Conference on Learning Representations},
  year={2023}
}

@article{scott2025differentially,
  title={Differentially Private Federated $ k $-Means Clustering with Server-Side Data},
  author={Scott, Jonathan and Lampert, Christoph H and Saulpic, David},
  journal={arXiv preprint arXiv:2506.05408},
  year={2025}
}

@inproceedings{chen2024one,
  title={One-Shot Secure Federated K-Means Clustering Based on Density Cores},
  author={Chen, Long and Zhao, Jing and Fan, Jian-Wei and Chen, Haonan and Zhao, Ziming and Wang, Guisheng and Wang, Chuang},
  booktitle={2024 International Joint Conference on Neural Networks (IJCNN)},
  pages={1--8},
  year={2024},
  organization={IEEE}
}

@article{zhang2024enhancing,
  title={Enhancing Group Fairness in Federated Learning through Personalization},
  author={Zhang, Yifan and Wang, Xiaofeng and Shen, Sicheng and Wang, Yunlong and Cheng, Long},
  journal={arXiv preprint arXiv:2407.19331},
  year={2024}
}

@article{xia2020distributed,
  title={Distributed K-Means clustering guaranteeing local differential privacy},
  author={Xia, Chang and Hua, Jingyu and Tong, Wei and Zhong, Sheng},
  journal={Computers \& Security},
  volume={90},
  pages={101699},
  year={2020},
  publisher={Elsevier}
}

@article{demelius2025recent,
  title={Recent advances of differential privacy in centralized deep learning: A systematic survey},
  author={Demelius, Lea and Kern, Roman and Tr{\"u}gler, Andreas},
  journal={ACM Computing Surveys},
  volume={57},
  number={6},
  pages={1--28},
  year={2025},
  publisher={ACM New York, NY}
}

@article{steinley2004properties,
  title={Properties of the hubert-arable adjusted rand index.},
  author={Steinley, Douglas},
  journal={Psychological methods},
  volume={9},
  number={3},
  pages={386},
  year={2004},
  publisher={American Psychological Association}
}

@article{mcdaid2011normalized,
  title={Normalized mutual information to evaluate overlapping community finding algorithms},
  author={McDaid, Aaron F and Greene, Derek and Hurley, Neil},
  journal={arXiv preprint arXiv:1110.2515},
  year={2011}
}

@misc{long2025,
      title={PA-CFL: Privacy-Adaptive Clustered Federated Learning for Transformer-Based Sales Forecasting on Heterogeneous Retail Data}, 
      author={Yunbo Long and Liming Xu and Ge Zheng and Alexandra Brintrup},
      year={2025},
      eprint={2503.12220},
      archivePrefix={arXiv},
      primaryClass={cs.LG},
      url={https://arxiv.org/abs/2503.12220}, 
}

\newpage

\appendix

\section*{Theoretical Analysis: Influence of Privacy Budget $\epsilon$ on Global Centroid Accuracy}

This analysis details how the accuracy of global centroids computed by the Global Federated Clustering (GFC) algorithm is influenced by the privacy budget, $\epsilon$, used in the client-side differential privacy mechanism. We first establish the sources of error, derive a general error bound, and then demonstrate how GFC's server-side approach mitigates the dominant noise term to achieve a tighter, more practical bound.

\subsection*{Step 1: Decomposing Privacy-Induced Error}

The server-side algorithm reconstructs a global data landscape using a set of local summaries $\{(\mathbf{c}_i, w_i)\}_{i \in \mathcal{S}}$ submitted by clients. Each summary consists of a local centroid position $\mathbf{c}_i$ and a corresponding mass $w_i$. The client-side privacy mechanism, which adds Laplace noise to individual data points $\mathbf{x}$ to create noisy points $\tilde{\mathbf{x}} = \mathbf{x} + \boldsymbol{\eta}$, introduces two distinct forms of error into these summaries:

\begin{enumerate}
    \item \textbf{Positional Perturbation ($\boldsymbol{\delta}_{C}$):} The noise added to data points causes the computed position of a local centroid, $\mathbf{c}_i$, to shift away from its true (non-private) position, $\mathbf{c}_{i, \text{true}}$. We define this shift as $\boldsymbol{\delta}_{C,i} = \mathbf{c}_i - \mathbf{c}_{i, \text{true}}$.

    \item \textbf{Mass Perturbation ($\delta_{W}$):} The mass $w_i$ is derived from the local cluster's inertia (the sum of squared intra-cluster distances). The noise $\boldsymbol{\eta}$ inflates these distances, perturbing the calculated mass from its true value, $w_{i, \text{true}}$. We define this scalar perturbation as $\delta_{W,i} = w_i - w_{i, \text{true}}$.
\end{enumerate}

Our goal is to bound the expected error in the final global centroid, $\mathbb{E}[\|\mathbf{c}_k^* - \mathbf{c'}_k^*\|]$, by analyzing how these two local error sources propagate through the GFC algorithm.

\subsection*{Step 2: Bounding the Positional Perturbation ($\boldsymbol{\delta}_{C}$)}

A local centroid $\mathbf{c}_i$ is the mean of $N_i$ noisy data points within its cluster. The positional perturbation $\boldsymbol{\delta}_{C,i}$ is the difference between the mean of the noisy points and the mean of the true points:
$$
\boldsymbol{\delta}_{C,i} = \left\| \frac{1}{N_i} \sum_{j=1}^{N_i} \tilde{\mathbf{x}}_j - \frac{1}{N_i} \sum_{j=1}^{N_i} \mathbf{x}_j \right\| = \left\| \frac{1}{N_i} \sum_{j=1}^{N_i} \boldsymbol{\eta}_j \right\|
$$
The noise for each of the $d$ components of $\boldsymbol{\eta}_j$ is drawn from a Laplace distribution, $\text{Laplace}(0, 1/\epsilon)$, which has a variance of $2/\epsilon^2$. The expected squared norm of a single noise vector is the sum of the variances across all dimensions:
$$
\mathbb{E}[\|\boldsymbol{\eta}_j\|^2] = \sum_{l=1}^d \text{Var}(\eta_{jl}) = \frac{2d}{\epsilon^2}
$$
The expected norm is therefore $\mathbb{E}[\|\boldsymbol{\eta}_j\|] = \mathcal{O}(\sqrt{d}/\epsilon)$. The perturbation $\boldsymbol{\delta}_{C,i}$ is the norm of the average of $N_i$ such noise vectors. By the principles of standard error, its expected magnitude scales as:
$$
\mathbb{E}[\|\boldsymbol{\delta}_{C,i}\|] \approx \frac{\mathbb{E}[\|\boldsymbol{\eta}_j\|]}{\sqrt{N_i}} = \mathcal{O}\left(\frac{\sqrt{d}}{\epsilon \sqrt{N_i}}\right)
$$
For simplicity, we consider the dominant relationship with the privacy budget, yielding an expected positional perturbation that is linear in $1/\epsilon$:
$$
\mathbb{E}[\|\boldsymbol{\delta}_{C,i}\|] = \mathcal{O}\left(\frac{1}{\epsilon}\right)
$$

\subsection*{Step 3: Bounding the Mass Perturbation ($\delta_{W}$)}

The mass $w_i$ is defined as $(I_i + \delta)^{-1}$, where $I_i$ is the cluster inertia. The noise $\boldsymbol{\eta}$ increases the expected inertia. The expected increase in squared distance for a single point is $\mathbb{E}[\|\boldsymbol{\eta}_j\|^2] = 2d/\epsilon^2$. Summing over the $N_i$ points in the cluster, the total expected increase in inertia, $\mathbb{E}[\Delta_I]$, is:
$$
\mathbb{E}[\Delta_I] \approx \sum_{j=1}^{N_i} \mathbb{E}[\|\boldsymbol{\eta}_j\|^2] = N_i \cdot \frac{2d}{\epsilon^2} = \mathcal{O}\left(\frac{N_i d}{\epsilon^2}\right)
$$
The resulting perturbation in mass, $\Delta_W = |w_{\text{noisy}} - w_{\text{true}}|$, can be approximated by a first-order Taylor expansion:
$$
\Delta_W \approx \left| \frac{d}{dI} (I_{\text{true}} + \delta)^{-1} \cdot \Delta_I \right| = (w_{\text{true}})^2 \cdot \Delta_I
$$
Taking the expectation, the mass perturbation is dominated by the variance of the noise, which is quadratic in $1/\epsilon$:
$$
\mathbb{E}[|\delta_{W,i}|] \approx (w_{\text{true}})^2 \cdot \mathbb{E}[\Delta_I] = \mathcal{O}\left(\frac{1}{\epsilon^2}\right)
$$

\subsection*{Step 4: The GFC Advantage via Structural Filtering}

A naive aggregation method, such as a direct weighted average of the local summaries, would be highly susceptible to the mass perturbation. The final error would be dominated by the term with the highest variance, leading to a pessimistic bound of $\mathcal{O}(1/\epsilon^2)$.

However, GFC employs a fundamentally different server-side aggregation based on identifying persistent structural features of the data landscape, which alters how these errors propagate. This approach relies on the following assumptions for practical scenarios:
\begin{enumerate}
    \item \textbf{Well-Separated Global Clusters:} The underlying data distribution contains distinct clusters.
    \item \textbf{Sufficient Data Aggregation:} The server receives a large number of local summaries, allowing statistical effects to emerge.
    \item \textbf{Unstructured Noise Impact:} The client-side noise is stochastic and does not systematically create or destroy large-scale data structures.
\end{enumerate}
GFC constructs a potential field $E(\mathbf{r}) = \sum_i w_i / (\|\mathbf{r} - \mathbf{c}_i\|^p + \delta)$ from the noisy summaries. The global centroids are identified as the most persistent peaks of this field. We analyze the impact of each error type on this field:
\begin{itemize}
    \item \textbf{Impact of Positional Perturbation:} The $\mathcal{O}(1/\epsilon)$ positional errors $\boldsymbol{\delta}_{C,i}$ cause the locations of the point masses to jitter. Due to the law of large numbers (per Assumption 2), the collective center of a cloud of jittered points remains very close to the true center. This introduces a small, well-behaved spatial shift in the location of the potential field's peaks. The error this introduces into the final global centroid's position is proportional to the average positional error, scaling as $\mathcal{O}(1/\epsilon)$.
    \item \textbf{Impact of Mass Perturbation:} The $\mathcal{O}(1/\epsilon^2)$ mass errors $\delta_{W,i}$ do not systematically shift the \textit{location} of the point cloud. Instead, they perturb the \textit{height} of the potential field at each point, introducing high-frequency, large-amplitude "bumpiness" onto the energy landscape. The server's filtration process is explicitly designed to be robust to this type of noise. It functions as a low-pass filter, identifying the large-scale, stable peaks (the "hills") while remaining insensitive to the fine-grained, high-variance noise (the "bumps on the hills"). This structural analysis effectively filters out the influence of the mass perturbation when determining the \textit{location} of a global centroid.
\end{itemize}

\subsection*{Step 5: The Effective Error Bound for GFC}

In the GFC framework, the final error in a global centroid's position, $\|\mathbf{c}_k^* - \mathbf{c'}_k^*\|$, is dominated by the component of noise that causes a systematic spatial shift, not the component that adds unstructured amplitude fluctuations.
\begin{enumerate}
    \item The high-variance mass perturbation, which scales as $\mathcal{O}(1/\epsilon^2)$, is effectively mitigated by the structural filtering step and does not significantly impact the \textit{location} of the computed centroid.
    \item The final error is therefore determined by the lower-variance positional perturbation, which causes a collective spatial shift that scales as $\mathcal{O}(1/\epsilon)$.
\end{enumerate}
Thus, for practical scenarios satisfying our assumptions, the GFC algorithm circumvents the worst-case scenario. The effective error bound for a global centroid's accuracy is determined by the well-behaved positional error term:
$$
\mathbb{E}[\|\mathbf{c}_k^* - \mathbf{c'}_k^*\|] = \mathcal{O}\left(\frac{1}{\epsilon}\right)
$$
This tighter bound is a direct consequence of GFC's novel server-side aggregation method, which intelligently separates and filters the different types of noise introduced by the client-side privacy mechanism.

\begin{figure*}[t]
    \centering
    \begin{minipage}[t]{0.32\textwidth}
        \centering
        \includegraphics[width=\linewidth]{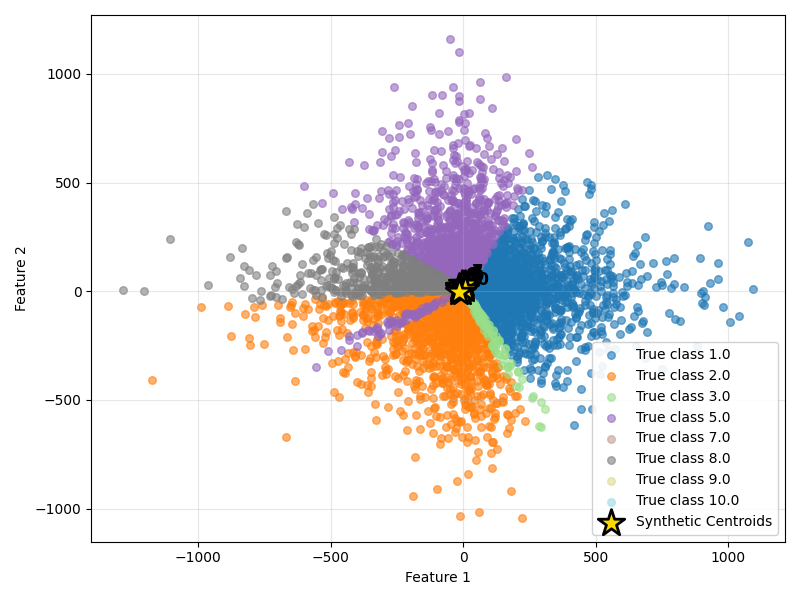}
        \par\vspace{0.1em}
        \textbf{(a)GFC}
    \end{minipage}
    \hfill
    \begin{minipage}[t]{0.32\textwidth}
        \centering
        \includegraphics[width=\linewidth]{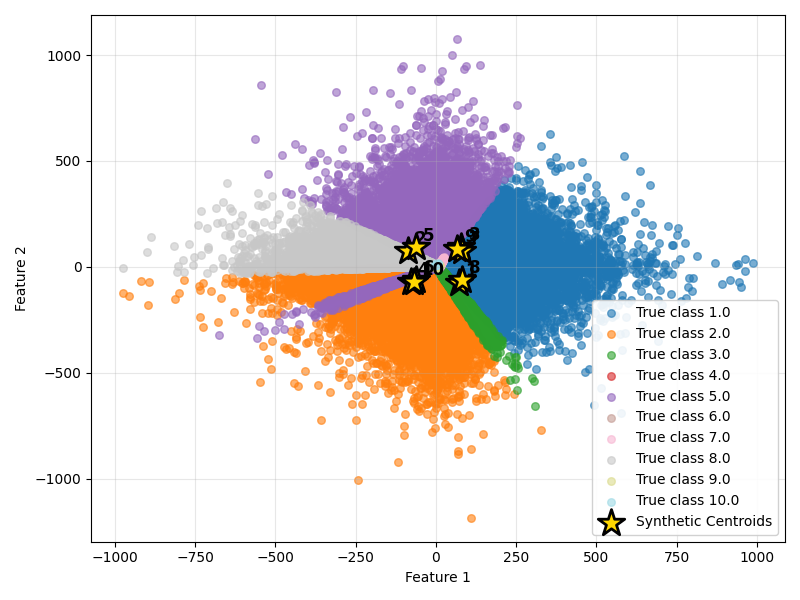}
        \par\vspace{0.1em}
        \textbf{(b)k-fed}
    \end{minipage}
    \hfill
    \begin{minipage}[t]{0.32\textwidth}
        \centering
        \includegraphics[width=\linewidth]{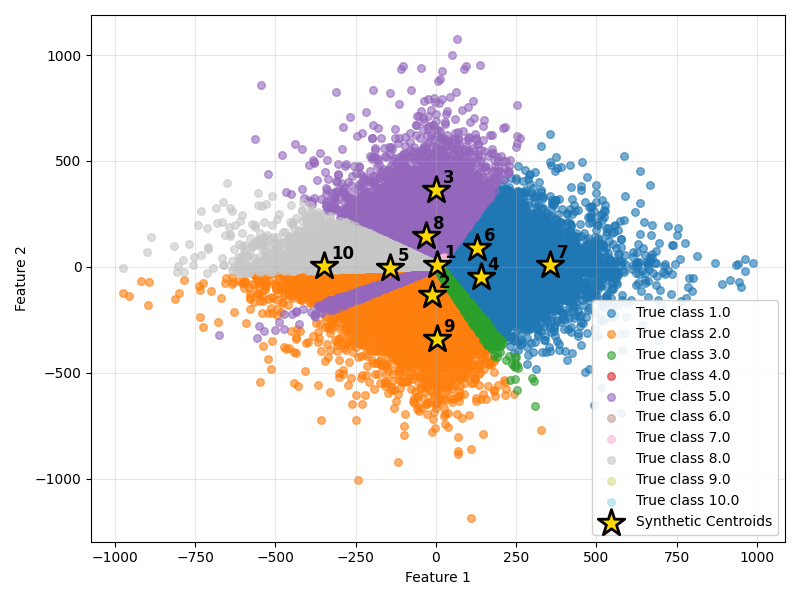}
        \par\vspace{0.1em}
        \textbf{(c)MUFC}
    \end{minipage}
    \caption{Topological Analysis for GFC on MNIST Data Visualized under 0.01 privacy budget via UMAP Projection}
    \label{fig:centroids_LDP}
\end{figure*}

\subsection{Detailed Algorithm}
\label{subsec:detailed_algorithm}

To provide a comprehensive understanding of our Gravitational Federated Clustering (GFC) framework, we decompose the overall procedure into three modular components: Client-Side Local Processing, Server-Side Gravitational Potential Field Construction, and Topological Persistence Clustering. Each component addresses a distinct aspect of the GFC pipeline.

\subsubsection*{Client-Side Local Processing}

The client-side algorithm (Algorithm~\ref{alg:client_phase}) transforms raw client data into privacy-preserving cluster representations that are transmitted to the server. This approach ensures that sensitive raw data never leaves the client devices.
GFC begins by applying Local Differential Privacy (LDP), adding Laplace noise scaled to $\Delta/\epsilon$ to each data point to ensure formal $\epsilon$-LDP guarantees at the client level (Line 2). It then performs K-means clustering on this privatized data to identify local patterns while strictly adhering to privacy constraints (Line 4). Finally, a quality-aware mass is calculated for each centroid, using intra-cluster variance to assign higher weights to more compact clusters; this mechanism enhances the algorithm's robustness against the noisy outcomes of local clustering (Lines 6-9).

\begin{algorithm}[t!]
\caption{Client-Side Local Processing}
\label{alg:client_phase}
\textbf{Input}: Local dataset $D_c$, Privacy budget $\epsilon$, Local cluster count $k$, Mass variance parameter $\sigma$. \\
\textbf{Output}: Set of weighted local centroids $\mathcal{S}_c = \{(C_{ci}, w_{ci})\}_{i=1}^{k}$.

\begin{algorithmic}[1]
\STATE Apply local differential privacy to the dataset:
    \STATE $\tilde{D}_c \gets D_c + \text{Lap}(0, \Delta/\epsilon)$
    \COMMENT{Add Laplace noise scaled by sensitivity $\Delta$ and privacy budget $\epsilon$}
\STATE Perform local clustering on the privatized data:
    \STATE $\{C_{c1},\ldots,C_{ck}\} \gets \text{KMeans}(\tilde{D}_c, k)$
    \COMMENT{Cluster privatized data into $k$ local centroids}
\STATE Calculate a quality-aware mass for each local centroid:
    \FOR{each centroid $C_{ci}$}
        \STATE $\text{variance}_{ci} \gets \frac{1}{|C_{ci}|} \sum_{x \in C_{ci}} \|x - C_{ci}\|^2$ 
        \COMMENT{Compute intra-cluster variance as compactness measure}
        \STATE $w_{ci} \gets \exp\left(-\frac{\text{variance}_{ci}}{2\sigma^2}\right)$ 
        \COMMENT{Higher mass for more compact, reliable clusters}
    \ENDFOR
\STATE \textbf{return} $\mathcal{S}_c \gets \{(C_{ci}, w_{ci})\}_{i=1}^{k}$ 
\COMMENT{Upload weighted centroids to server}
\end{algorithmic}
\end{algorithm}

\subsubsection*{Server-Side Gravitational Potential Field Construction}

The server aggregates all client contributions and constructs a continuous potential field (Algorithm~\ref{alg:server_potential}) that represents the underlying data distribution without accessing raw client data. This process begins by generating a synthetic grid that uniformly samples the data space, where the sampling density is controlled by the parameter $\alpha$ to create a canvas for potential field measurement (Lines 2-4). The gravitational potential is then calculated by treating each local centroid as a mass particle, with the softening factor $\delta$ preventing numerical instability and creating a smoothed density estimate (Lines 6-10). The resulting potential field $E(G)$ effectively captures the intrinsic cluster structure while maintaining robustness to noise and client heterogeneity.

\begin{algorithm}[t!]
\caption{Server-Side Potential Field Construction}
\label{alg:server_potential}
\textbf{Input}: Aggregated weighted centroids $\mathcal{S} = \bigcup_{c=1}^C \mathcal{S}_c$, Synthetic data multiplier $\alpha$, Softening factor $\delta$, Data bounds $\mathcal{B}$. \\
\textbf{Output}: Synthetic grid $G$, Potential energy $E(g)$ for all $g \in G$.
\begin{algorithmic}[1]
\STATE Generate a synthetic grid to probe the potential field:
    \STATE $N_{\text{synthetic}} \gets \alpha \cdot |\mathcal{S}|$ 
    \COMMENT{Scale synthetic points by total centroids count}
    \STATE $G \gets \text{UniformSample}(\mathcal{B}, N_{\text{synthetic}})$ 
    \COMMENT{Sample uniformly within aggregated data bounds}
\STATE Compute the gravitational potential energy at each synthetic point:
    \FOR{each synthetic point $g_j \in G$}
        \STATE $E(g_j) \gets 0$
        \FOR{each weighted centroid $(C_i, w_i) \in \mathcal{S}$}
            \STATE $E(g_j) \gets E(g_j) + \frac{w_i}{\|g_j - C_i\|^2 + \delta}$ 
            \COMMENT{Accumulate gravitational influence from all centroids}
        \ENDFOR
    \ENDFOR
\STATE \textbf{return} $G$, $E(G)$
\end{algorithmic}
\end{algorithm}

\subsubsection*{Topological Persistence Clustering on Server}

The final stage (Algorithm~\ref{alg:topological_clustering}) employs topological data analysis to extract stable cluster centers from the potential field by analyzing its persistence structure. This process begins with threshold generation, which creates a fine-grained sequence to capture the evolution of sub-level sets as energy increases (Lines 4-6). For computational efficiency, the threshold sequence can be optimized by sampling adaptively from midpoints between energy levels or using binary search strategies when the grid is large. The core of the method lies in merge tree construction, which tracks how connected components are born, merge, and persist across energy levels, providing crucial robustness to noise (Lines 8-19). Within this framework, energy-weighted centroid calculation ensures cluster centers position themselves in high-density regions of the potential field (Lines 11-12), while persistence-based selection prioritizes clusters that maintain structural stability across threshold ranges, effectively distinguishing true signal from noise artifacts (Lines 22-30).

\begin{algorithm}[t!]
\caption{Topological Persistence Clustering on Server}
\label{alg:topological_clustering}
\textbf{Input}: Grid points $G$, Energy field $E: G \to \mathbb{R}^+$, Target clusters $n_c$, Connectivity radius $r$ \\
\textbf{Output}: Centroids $F = \{f_1, \dots, f_{n_c}\}$

\begin{algorithmic}[1]
\STATE Initialize merge tree $\mathcal{T} \gets \emptyset$
\STATE Generate threshold sequence $H = \{h_1 < h_2 < \cdots < h_m\}$ from sorted $E(G)$
    \COMMENT{Can be optimized via adaptive sampling or binary search}

\FOR{$h \in H$}
    \STATE Compute sublevel set: $S_h = \{g \in G \mid E(g) \leq h\}$
    \STATE Extract components: $\mathcal{C}_h = \pi_0(S_h, r)$ \COMMENT{0-th homology with radius $r$}
    
    \IF{$\mathcal{C}_h \neq \mathcal{C}_{h-\epsilon}$}
        \STATE $\mathcal{T} \leftarrow \mathcal{T} \cup \{(C, h) \mid C \in \mathcal{C}_h \setminus \mathcal{C}_{h-\epsilon}\}$
        \FOR{each newborn component $C \in \mathcal{C}_h$}
            \STATE Compute centroid: $\mu_C = \dfrac{\sum_{g \in C} E(g) \cdot g}{\sum_{g \in C} E(g)}$
            \STATE $\mathcal{T} \leftarrow \mathcal{T} \cup \{(C, \mu_C, h)\}$
        \ENDFOR
    \ENDIF
    
    \IF{$|\mathcal{T}_{\text{active}}| \geq n_c$}  
        \STATE \textbf{break} \COMMENT{Early termination when sufficient candidates found}
    \ENDIF
\ENDFOR

\STATE Rank candidates by persistence: $\mathcal{P} = \{(C, \text{persist}(C)) \mid C \in \mathcal{T}\}$
\STATE $F \gets \{\mu_C \mid C \in \text{top}_{n_c}(\mathcal{P})\}$

\IF{$|F| < n_c$}
    \STATE $F \gets F \cup \{\mu_C \mid C \in \text{top}_{n_c-|F|}(\mathcal{T}_{\text{energy}})\}$
    \IF{$|F| < n_c$}
        \STATE $F \gets F \cup \{g \in \text{top}_{n_c-|F|}(E(G))\}$
    \ENDIF
\ENDIF

\STATE \textbf{return} $F$
\end{algorithmic}
\end{algorithm}

The three-algorithm decomposition enables clear separation of concerns, with privacy-sensitive operations confined to client devices, gravitational modeling for robust aggregation, and topological methods for stable cluster extraction. This modular design facilitates both theoretical analysis and practical implementation.

\section{Non-IID Dataset}
\label{app:noniid}

Our evaluation framework incorporates diverse datasets spanning synthetic waveforms, biological measurements, motion patterns, and medical diagnostics. The synthetic \textit{Waveform} dataset (5,000 samples) contains three distinct waveform patterns, while biological datasets include \textit{Abalone} (4,177 marine specimens) and \textit{Seeds} (210 grain varieties). Motion analysis is represented by \textit{Gesture} (1,747 samples of hand movements), and medical applications feature \textit{Heart} (303 cardiac cases), \textit{Breast} (277 tumor classifications), and \textit{Thyroid} (215 endocrine evaluations). Smaller datasets like \textit{Balanced Scale} (625 simulated measurements) provide additional testing scenarios. This heterogeneous collection enables comprehensive evaluation across different data domains and scales.

The federated scenario is constructed by partitioning the dataset into \textit{num\_clients} subsets with non-IID label distributions. First, $k$-means clustering is applied to group the data based on similarity. Then, for each client assignment, two random $k$-means clusters are selected, and $\min(r_1, r_2, r_3)$ points are allocated to the current client, where $r_1$, $r_2$, and $r_3$ represent the available data counts from the selected clusters and the remaining capacity of the client. This process is repeated for the first \textit{num\_clients}$- 1$ clients, while all remaining points are assigned to the final client. This strategy ensures heterogeneous label distributions across clients while maintaining a balanced representation of clusters.

For non-IID partitioning, as shown in Algorithm \ref{alg:noniid_gen}, we implement a cluster-based distribution mechanism, which following the settings from teh MUFC\citep{PanSima:Unlearning}. The process begins with k-means clustering (where $k$ matches the ground truth class count) to identify natural data groupings. Each client except the last receives points from two randomly selected clusters, with the allocation quantity determined by three constraints: (1) a random size parameter, (2) the remaining cluster capacity, and (3) available unassigned points. The final client collects all residual data points, ensuring complete distribution while maintaining non-IID characteristics. This approach generates realistic data skewness while preventing extreme allocation imbalances. Following the federated clustering dataset settings in \citet{wang2024one}, we partition each dataset uniformly across 10 clients with non-IID distributions. For the small to medium datasets, we keep their original dimensions. For large-scale datasets, we first apply dimension reduction to 2D representations (using UMAP shown in Figure \ref{fig:centroids_LDP}) to optimize communication efficiency during local client operations. These preprocessed, non-IID distributed embeddings then serve as input to the federated clustering pipeline. This approach maintains the privacy-preserving benefits of federated learning while significantly reducing bandwidth requirements - particularly crucial for large datasets.

\begin{algorithm}[ht]
\caption{Non-IID Client Data Generation}
\label{alg:noniid_gen}
\textbf{Input}: Raw dataset $D$, number of clients $num\_clients$\\
\textbf{Output}: Non-IID partitioned data of clients

\begin{algorithmic}[1]
\STATE $n\_clusters \gets$ number of classes in $D$ (from ground truth)
\STATE $size \gets |D| / num\_clients$
\STATE $centers, assignment \gets$ K-means($D$, $n\_clusters$)
\FOR{$i = 1$ to $num\_clients-1$}
    \STATE Randomly select two clusters $(s_1, s_2)$ from $assignment$
    \FOR{$s \in \{s_1, s_2\}$}
        \STATE $r_1 \gets$ random integer in $[size/(n\_clusters/2), size]$
        \STATE $r_2 \gets size - |s|$
        \STATE $r_3 \gets$ unassigned points in $s$
        \STATE Assign $min(r_1, r_2, r_3)$ points from $s$ to client $c_i$
    \ENDFOR
\ENDFOR
\STATE Assign all remaining points to client $c_{num\_clients}$
\STATE \textbf{return} client partitions
\end{algorithmic}
\end{algorithm}

\begin{figure}[h]
    \centering
    \begin{subfigure}[b]{0.23\textwidth}
        \centering
        \includegraphics[width=\linewidth]{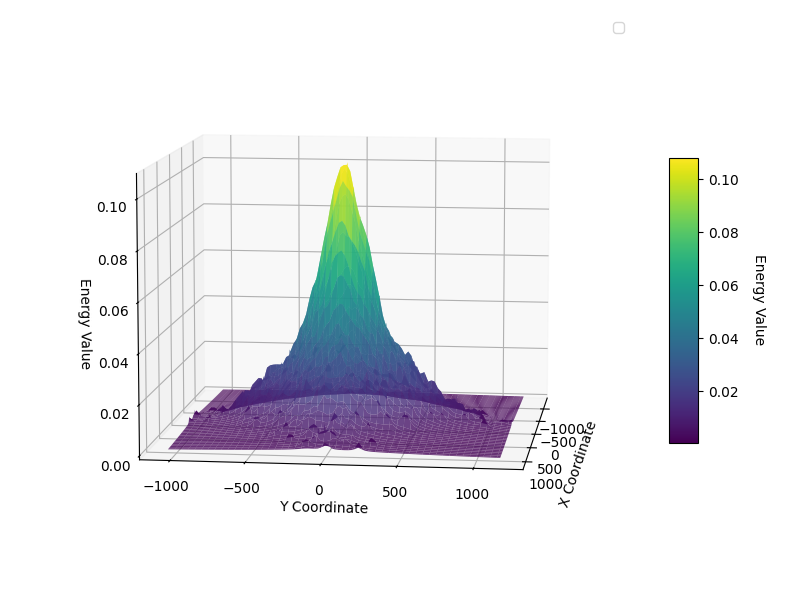}
        \caption{Energy under strong LDP}
        \label{fig:energy}
    \end{subfigure}
    \hfill
    \begin{subfigure}[b]{0.23\textwidth}
        \centering
        \includegraphics[width=\linewidth]{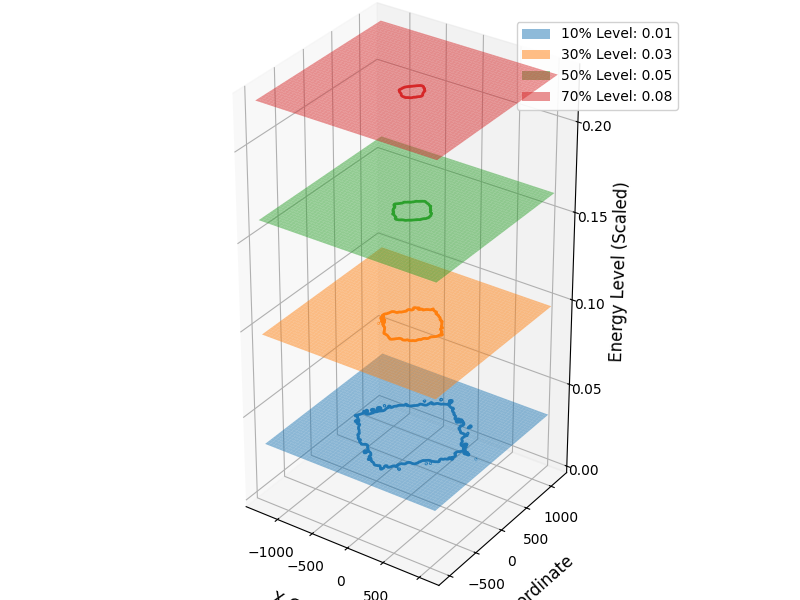}
        \caption{Topologies filter}
        \label{fig:topology}
    \end{subfigure}
    \caption{GFC's topological analysis on MNIST data ($\epsilon=0.01$).}
    \label{fig:low_privacy}
\end{figure}

\begin{figure*}[t]
    \centering
    \begin{minipage}[t]{0.32\textwidth}
        \centering
        \includegraphics[width=\linewidth]{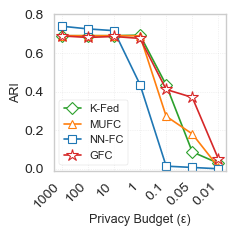}
        \par\vspace{0.1em}
        \textbf{(a) Seeds \textsubscript{ARI}}
    \end{minipage}
    \hfill
    \begin{minipage}[t]{0.32\textwidth}
        \centering
        \includegraphics[width=\linewidth]{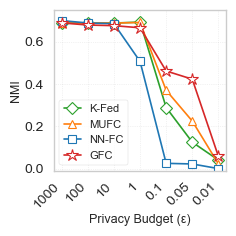}
        \par\vspace{0.1em}
        \textbf{(b) Seeds\textsubscript{NMI}}
    \end{minipage}
    \hfill
    \begin{minipage}[t]{0.32\textwidth}
        \centering
        \includegraphics[width=\linewidth]{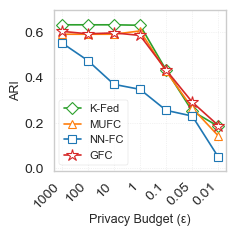}
        \par\vspace{0.1em}
        \textbf{(c) Thyroid \textsubscript{ARI}}
    \end{minipage}
    \hfill
    \begin{minipage}[t]{0.32\textwidth}
        \centering
        \includegraphics[width=\linewidth]{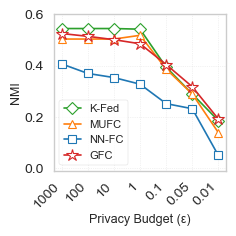}
        \par\vspace{0.1em}
        \textbf{(d) Thyroid\textsubscript{NMI}}
    \end{minipage}
    \begin{minipage}[t]{0.32\textwidth}
        \centering
        \includegraphics[width=\linewidth]{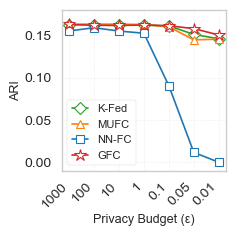}
        \par\vspace{0.1em}
        \textbf{(e) Celltype \textsubscript{ARI}}
    \end{minipage}
    \hfill
    \begin{minipage}[t]{0.32\textwidth}
        \includegraphics[width=\linewidth]{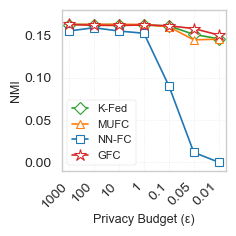}
        \par\vspace{0.1em}
        \textbf{(f) Celltype\textsubscript{NMI}}
    \end{minipage}
    
    \caption{Performance comparison of GFC, NN-FC, K-Fed, and MUFC across varying privacy budgets $\varepsilon$ (smaller $\varepsilon$ denotes stronger privacy). Mean ARI scores are shown in (a, c, e, g), while corresponding mean NMI scores are in (b, d, f, h). GFC results are highlighted in red.}
    \label{fig:varying_DP2}
\end{figure*}

\section{Implementation}
\label{app:Implementation}

\paragraph{\textbf{Hyperparameters}}
We systematically optimize all four models across varying privacy budgets ($\epsilon$). For NN-FC (Nearest Neighbor Federated Clustering), we grid-search $k_1$ (local neighbors) over ${10, 20, 50, 100, 200, 400}$ and $k_2$ (global neighbors) over ${10, 20, 50}$, while fixing $k_3$ (final clusters) to the ground-truth class count. K-Fed requires tuning $k$ (server-side K-means clusters) to match the dataset’s true classes, while $k'$ (uploaded centers per client) is swept from $1$ to the dataset-specific upper bound (determined by K-means convergence). Similarly, MUFC fixes its server-side $k$ to the true cluster count and explores $k' \in {1, 2, \dots, k_{\text{max}}}$, where $k_{\text{max}}$ is derived from per-dataset stability constraints. 
For GFC, we adapt hyperparameters based on dataset scale requirements. The number of local centroids per client is set to $k \in {1, 2, 5, 10}$ for small-to-medium datasets and ${100, 200, 400}$ for larger datasets. We evaluate gravitational softening parameters $\delta \in {5\times10^{-4}, 0.5, 500}$ to balance cluster separation and merging. The synthetic data generation is controlled through multiplier $\alpha \in {1, 2, 10}$, while neighborhood radius $r$ is dynamically determined as the 1st percentile of non-zero inter-point distances ($r = \text{percentile}(d_{>0}, 1)$). Neighborhood counts are explored in the range 10 to optimize local density estimation.

We also propose a heuristic tuning strategy to dynamically adapts to privacy constraints through a continuum between two operational regimes: When dealing with low-$\epsilon$ (noisy) data where cluster boundaries are ambiguous, we employ strong gravitational smoothing ($\delta=400$) with fixed neighborhood radius ($r=1$) to force global centroid consolidation, focusing optimization exclusively on the number of local centroids ($k$) and synthetic data multiplier ($\alpha$) to overcome noise-induced fragmentation. Conversely, for high-$\epsilon$ (clean) data where original structures are preserved, we switch to weak smoothing ($\delta=5\times10^{-4}$) and activate full parameter optimization (including $r$ and neighborhood counts 10) to resolve fine-grained regional patterns. The transition between these regimes is automatically governed by the privacy budget $\epsilon$, creating a smooth interpolation where intermediate values blend both strategies - maintaining centroid cohesion in noisy regions while permitting local structure emergence in well-defined areas, all without compromising computational efficiency.

\paragraph{\textbf{Computational Resources}}
All experiments were conducted on a high-performance computing cluster with specific hardware and software configurations. The operating system used was Ubuntu 20.04.6 LTS with a Linux kernel version of 5.15.0-113-generic. The CPU was an Intel(R) Xeon(R) Platinum 8368 processor running at 2.40 GHz, and the GPU was an NVIDIA GeForce RTX 4090 with CUDA support for accelerated deep learning computations. The software stack included Python 3.8, PyTorch 1.12, and TensorFlow 2.10 for model implementation and training.

\section{Centroids selection under strong LDP}
\label{app:algo}

Under strong local differential privacy ($\epsilon = 0.01$), the Laplace noise transforms the true data distribution $p_{\text{true}}(x)$ into a privatized version $p_{\text{DP}}(x) = p_{\text{true}}(x) \ast \text{Laplace}(0,\beta)$ where $\beta = \Delta/\epsilon$. 
This operation preserves the distribution's modes while causing them to merge into overlapping peaks, as shown by the key relationship $\arg\max_x p_{\text{DP}}(x) = \arg\max_x p_{\text{true}}(x) + \mathcal{O}(\beta)$. 

The $L_1$ error bound $\|\hat{\mu}_k - \mu_k\|_{1} \leq \frac{2\Delta}{\epsilon n_k}$ guarantees that centroids estimated from $p_{\text{DP}}$ remain near high-density regions, explaining GFC's superior clustering performance (ARI/NMI) over k-FED and MUFC. Sparse centroids fail because they violate this bound, suffering from noise amplification
$\|\hat{\mu} - \mu_{\text{true}}\|_1 \geq \beta/\sqrt{p_{\text{DP}}(\hat{\mu})}$.
Figures \ref{fig:low_privacy} and \ref{fig:topology} demonstrate that GFC can identify overlapping high-energy regions  and then recover the distribution of close centroids in high-energy regions by adaptively tuning the number of synthetic data points and the parameter $\delta$ (for $\epsilon=0.01$), ensuring compliance with theoretical bounds. In contrast, MUFC and k-Fed struggle to adjust centroids flexibly in high-energy regions under strong Local Differential Privacy (LDP) constraints, leading to suboptimal performance.

\section{Clustering performance under varying privacy budgets}

We also evaluate clustering quality (ARI/NMI) across a wide privacy spectrum ($\epsilon = 1000$ to $0.01$) for the rest of datasets spanning different scales: \textit{Seeds} (small), \textit{Thyroid} (medium), and \textit{Celltype} (large). Figure~\ref{fig:varying_DP2} reveals three key findings:
For the large-scale Celltype dataset with moderate privacy constraints ($\epsilon \geq 0.1$), GFC achieves comparable performance to both K-Fed and MUFC. However, GFC becomes superior when privacy constraints tighten ($\epsilon < 0.1$), demonstrating better noise robustness. Besides, similar trends emerge for \textit{Seeds} and \textit{Thyroid}. GFC remains competitive with baselines at moderate privacy levels ($1000 \geq \epsilon \geq 0.1$) but consistently outperforms them under strict privacy ($\epsilon \leq 0.1$). The advantage of GFC grows as $\epsilon$ decreases, highlighting its effectiveness for high-privacy regimes where other methods degrade. This pattern holds consistently across all dataset scales.

\section{Sensitivity to local centroids number}
\label{app:delta}

We evaluate GFC's sensitivity to the number of uploaded local centroids under varying privacy budgets ($\epsilon$), with fixed gravitational parameters ($\delta$), synthetic data multiplier ($\alpha$), and neighborhood radius ($r$). As shown in Tables~\ref{tab:local_abalone} (Abalone) and~\ref{tab:local_mnist} (MNIST), two key patterns emerge: (1) Under strong privacy constraints ($\epsilon \ll 1$), clustering performance shows significant dependence on the number of local centroids, with optimal results typically achieved at intermediate values (15-25 centroids for Abalone, 100-400 for MNIST); (2) In low-privacy regimes ($\epsilon = 1000$), GFC demonstrates remarkable robustness to centroid count variations, maintaining stable ARI/NMI values across wide parameter ranges. This differential sensitivity reflects the noise-amplification effect in low-$\epsilon$ settings, where insufficient centroids may fail to capture the underlying distribution like in MNIST dataset while excessive centroids propagate noise like in Abalone dataset.

\begin{table}[t]
\centering

\begin{tabular}{cc|cccc}
\toprule
\multirow{2}{*}{$\epsilon$} & \multirow{2}{*}{Metric} & \multicolumn{4}{c}{Local Centroids Number} \\
\cmidrule(lr){3-6}
 & & 10 & 15 & 20 & 25 \\ 
\midrule
\multirow{2}{*}{1000} 
 & ARI & 0.1647 & \textbf{0.1886} & 0.1611 & 0.1611 \\
 & NMI & 0.1446 & \textbf{0.1617} & 0.1490 & 0.1490 \\ 
\bottomrule
\end{tabular}

\vspace{0.5cm}

\begin{tabular}{cc|cccc}
\toprule
\multirow{2}{*}{$\epsilon$} & \multirow{2}{*}{Metric} & \multicolumn{4}{c}{Local Centroids Number} \\
\cmidrule(lr){3-6}
 & & 1 & 5 & 10 & 15 \\ 
\midrule
\multirow{2}{*}{0.01} 
 & ARI & \bf{0.1615} & NA & NA & NA \\
 & NMI & \bf{0.1514} & NA & NA & NA\\ 
\bottomrule
\end{tabular}

\vspace{0.2cm}
\begin{minipage}{\linewidth}
\footnotesize
\textit{Note}: Bold values indicate the best ARI performance for each $\epsilon$ level. Results show the impact of varying local centroids number under different privacy constraints.
\end{minipage}
\caption{Impact of Local Centroids Number $k$ and Privacy Budget ($\epsilon$) on Clustering Performance (Abalone Dataset)}
\label{tab:local_abalone}
\end{table}

\begin{table}[t]
\centering

\begin{tabular}{cc|cccc}
\toprule
\multirow{2}{*}{$\epsilon$} & \multirow{2}{*}{Metric} & \multicolumn{4}{c}{Local Centroids Number} \\
\cmidrule(lr){3-6}
 & & 100 & 200 & 300 & 400 \\ 
\midrule
\multirow{2}{*}{1000} 
 & ARI & 0.7742 & 0.7537 & 0.8180 & \bf{0.8870} \\
 & NMI & 0.8489 & 0.8394 & 0.8678 & \bf{0.8980} \\ 
\bottomrule
\end{tabular}

\vspace{0.5cm}

\begin{tabular}{cc|cccc}
\toprule
\multirow{2}{*}{$\epsilon$} & \multirow{2}{*}{Metric} & \multicolumn{4}{c}{Local Centroids Number} \\
\cmidrule(lr){3-6}
 & & 100 & 200 & 300 & 400 \\ 
\midrule
\multirow{2}{*}{0.01} 
 & ARI & 0.0392 & 0.1690 & 0.3410 & \bf{0.5626} \\
 & NMI & 0.3493 & 0.3874 & 0.5537 &\bf{ 0.7416} \\ 
\bottomrule
\end{tabular}

\vspace{0.2cm}
\begin{minipage}{\linewidth}
\footnotesize
\textit{Note}: Bold values indicate the best ARI performance for each $\epsilon$ level. Results show the impact of varying local centroids number under different privacy constraints.
\end{minipage}
\caption{Impact of Local Centroids Number $k$ and Privacy Budget ($\epsilon$) on Clustering Performance (MNIST Dataset)}
\label{tab:local_mnist}
\end{table}

\section{Federated Clustering Efficiency}
\label{app:efficiency}

\begin{figure}[h]
    \centering
    \begin{subfigure}[b]{\columnwidth}
        \centering
        \includegraphics[width=\linewidth]{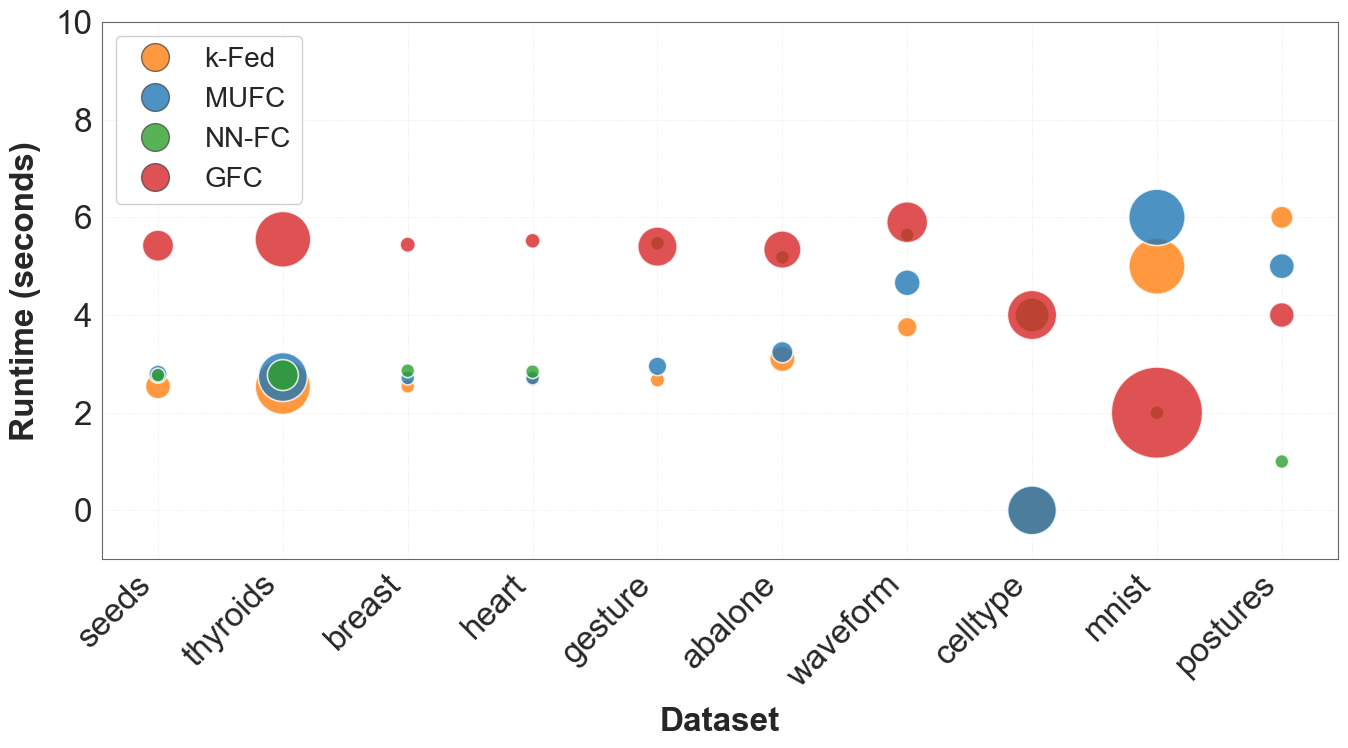}
        \caption{ARI}
        \label{fig:time_a}
    \end{subfigure}
    \caption{Adjusted Rand Index (ARI) scores and Normal}
    \label{fig:efficiency}
\end{figure}

We evaluate the computational efficiency and clustering accuracy (ARI) across four federated clustering methods under strict privacy constraints ($\epsilon=0.01$) with 10 clients, testing on 10 benchmark datasets (with large datasets scaled down by 1/20 for visualization). 
As shown in Figure \ref{fig:time_a}, results demonstrate that GFC achieves superior accuracy improvements (150\%+) compared to baselines, particularly on large datasets where its gravitational field computation and synthetic data generation ($n$) enhance cluster quality despite increased computational overhead. 
To further enhance computational efficiency, we introduce an alternative approach using image-based labels (nlabel) to replace the traditional adjacency neighborhood calculation in topological analysis. 
Besides, for small-to-medium datasets, GFC maintains competitive efficiency, matching NN-FC's runtime while approaching the speed of k-Fed and MUFC in many cases. Combined with scalability results in Table~\ref{tab:testing_charges}, these findings highlight GFC's flexible trade-off between accuracy and efficiency through strategic adjustment of synthetic data volume.

Considering the settings of the large dataset, they are reduced to 2-dimensional representations before federated clustering (a preprocessing step applied uniformly across all models), We develop an efficient topological analysis pipeline that replaces conventional adjacency calculations with image processing techniques: we first convert the 2D projection into a W×H grayscale image and apply adaptive thresholding to identify cluster regions.
As demonstrated in \ref{fig:efficiency}, this optimized version of GFC not only preserves its accuracy advantages but actually surpasses both k-Fed and MUFC in computational efficiency when processing dimensionally-reduced data. 
This alternative method proves particularly effective for high-dimensional datasets where the topological structure can be preserved in lower-dimensional embeddings, enabling faster gravitational field computations without compromising the quality of synthetic data generation.

\section{Scalability}
\label{app:delta}

\begin{table}[h]
\centering
\begin{tabular}{lccccc}
\hline
\multirow{2}{*}{N} & \multirow{2}{*}{Metric} & \multicolumn{4}{c}{FC Methods} \\
\cline{3-6}
 & & k-Fed & MUFC & NN-FC & GFC (Ours) \\ \hline
100 & ARI & NA & NA & NA & \textbf{1.51} \\
    & NMI & NA & NA & NA & \textbf{4.21} \\ \hline
1000 & ARI & NA & 0.01 & NA & \textbf{0.13} \\
     & NMI & NA & 0.03 & NA & \textbf{0.52} \\ \hline
\end{tabular}
\vspace{0.2cm}
\begin{minipage}{\linewidth}
\footnotesize
\textit{Note}: NA indicates cases where the model failed to produce meaningful clustering (ARI/NMI = 0). Bold values highlight superior performance. Results demonstrate GFC's scalability advantage under strong privacy constraints ($\epsilon=0.01$).
\end{minipage}
\caption{Performance Comparison of Federated Clustering Models Under Strict Privacy Budget ($\epsilon=0.01$) Measured by ARI and NMI Metrics}
\label{tab:scalability_results}
\end{table}

To evaluate GFC's robustness in real-world deployment scenarios, we conduct large-scale experiments extending from the conventional 10-client setting (typical for institutional federated learning) to 100-1000 client configurations (representing edge computing environments), all under strong local differential privacy (LDP) with $\epsilon=0.01$. As demonstrated in Table~\ref{tab:scalability_results}, for the MNIST datasets, existing one-shot federated clustering methods (K-Fed, MUFC, NN-FC) completely fail (producing NA results) due to the compounded noise effects from numerous clients, GFC maintains remarkable clustering performance. This superior performance highlights GFC's unique capability to preserve meaningful cluster structures even when both the participant scale and privacy constraints push conventional approaches beyond their operational limits.

Our scalability analysis reveals a critical challenge in federated clustering: when fixing the total dataset size while increasing the number of clients (from 10 to 100-1000), each client's local data becomes increasingly sparse. This sparsity amplifies the relative impact of LDP noise ($\varepsilon$=0.01) because: (1) the signal-to-noise ratio deteriorates as meaningful local patterns become obscured by proportionally larger noise, and (2) with fewer samples per client, local centroids become less reliable. Traditional one-shot methods fail completely (producing NA results) because they cannot aggregate these noisy, fragmented views effectively. However, GFC succeeds through its energy-based approach that: (a) gradually refines clusters by detecting consensus patterns across clients, (b) uses gravitational smoothing to distinguish true signals from noise artifacts, and (c) adapts synthetic data generation to compensate for sparsity. This explains GFC's maintained performance (ARI=1.51/0.13, NMI=4.21/0.52 for 100/1000 clients) while others fail.

\end{document}